\documentclass{ecai} 

\usepackage{latexsym}
\usepackage{amssymb, amsmath, amsthm, amsfonts}
\usepackage{booktabs}
\usepackage{enumitem}
\usepackage{graphicx}
\usepackage{color}
\usepackage{balance}

\usepackage{url}
\usepackage[utf8]{inputenc}
\usepackage[small]{caption}
\usepackage[switch]{lineno}
\usepackage[ruled, lined, linesnumbered, commentsnumbered, longend]{algorithm2e}
\usepackage{multirow}
\usepackage{adjustbox}
\usepackage{subcaption}
\usepackage{microtype}
\usepackage{xcolor}
\usepackage{dirtytalk}

\newtheorem{definition}{Definition}
\newtheorem{theorem}{Theorem}
\newtheorem{lemma}{Lemma}
\newtheorem{remark}{Remark}
\newtheorem{corollary}{Corollary}

\newcommand{\LL}{\mathcal{L}}

\begin{document}


\begin{frontmatter}


\paperid{8968} 


\title{Domain Generalization via Pareto Optimal Gradient Matching}


\author[A]{\fnms{Khoi}~\snm{Do}
\footnote{A part of this work was done when Khoi Do was with Hanoi University of Science and Technology.}
\footnote{Equal contribution.}}
\author[E]{\fnms{Duong}~\snm{Nguyen}\footnotemark\footnote{A part of this work was done when Minh-Duong Nguyen was with Pusan National University.}}
\author[C]{\fnms{Nam-Khanh}~\snm{Le}}
\author[A]{\fnms{Quoc-Viet}~\snm{Pham}}
\author[A]{\fnms{Binh-Son}~\snm{Hua}}
\author[B]{\fnms{Won-Joo}~\snm{Hwang}\thanks{Corresponding Author. Email: wjhwang@pusan.ac.kr}}

\address[A]{School of Computer Science and Statistics, Trinity College Dublin, Dublin, Ireland}
\address[B]{Pusan National University, Busan, South Korea}
\address[C]{SoICT, Hanoi University of Science and Technology, Hanoi, Vietnam}
\address[E]{College of Engineering \& Computer Science, Vin University, Hanoi, Vietnam}


\begin{abstract}
In this study, we address the gradient-based domain generalization problem, where predictors aim for consistent gradient directions across different domains. Existing methods have two main challenges. First, minimization of gradient empirical distance or gradient inner products (GIP) leads to \emph{gradient fluctuations} among domains, thereby hindering straightforward learning. Second, the direct application of gradient learning to joint loss function can incur \emph{high computation overheads} due to second-order derivative approximation. To tackle these challenges, we propose a new Pareto Optimality Gradient Matching (POGM) method. In contrast to existing methods that add gradient matching as regularization, we leverage gradient trajectories as collected data and apply independent training at the meta-learner. In the meta-update, we maximize GIP while limiting the learned gradient from deviating too far from the empirical risk minimization gradient trajectory. By doing so, the aggregate gradient can incorporate knowledge from all domains without suffering gradient fluctuation towards any particular domain. Experimental evaluations on datasets from DomainBed demonstrate competitive results yielded by POGM against other baselines while achieving computational efficiency.
\end{abstract}

\end{frontmatter}


\section{Introduction}

Domain generalization (DG) has emerged as a significant research field in machine learning, owing to its practical relevance and parallels with human learning in new environments. In DG frameworks, learning occurs across multiple datasets collected from diverse environments, with no access to data from the target domain \cite{2023-DG-Survey1}. Various strategies have been proposed to address DG challenges, including distributional robustness \cite{2022-DG-QRM}, domain-invariant representations \cite{2018-DG-CIAN,2020-DG-EntropyReg,2021-DG-DomainSpecificFeatures}, invariant risk minimization \cite{2021-DG-IRL-DDT,2022-DG-KLGuided}, and data augmentation \cite{2022-DG-CMixup,2021-DG-MixStyle,2021-DG-FeatureAugmentation}. 

\begin{figure}[!ht]
    \centering
    \includegraphics[width = \linewidth]{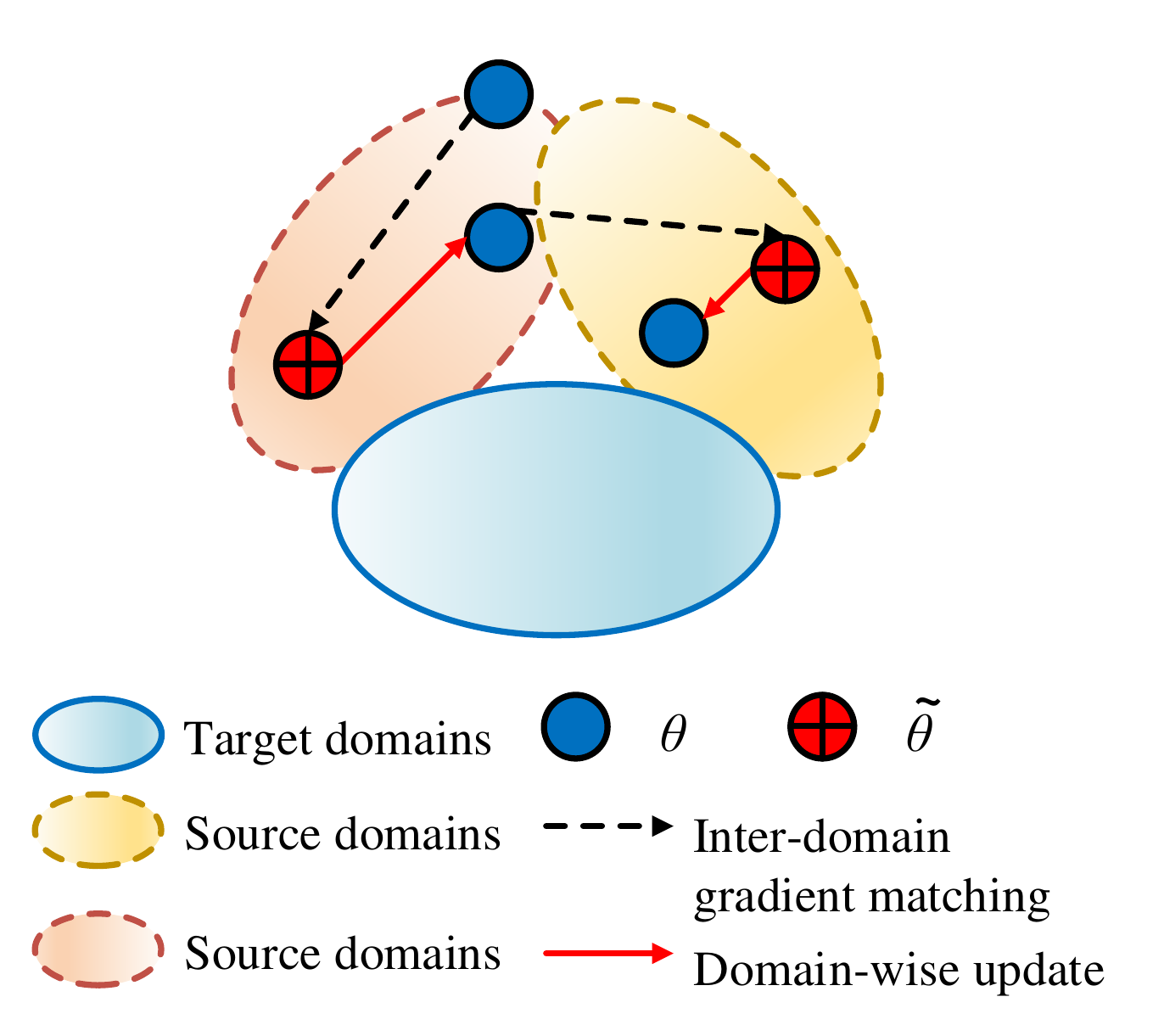} 
    \caption{Issues in current gradient-based DG methods. Fish introduces \emph{gradient fluctuations} across domains. Specifically, during each meta-loop update, the meta model $\theta$ is updated in the direction of the inner model $\widetilde{\theta}$ via a Reptile-based update $\theta\leftarrow\theta + \epsilon (\widetilde{\theta}-\theta)$. However, when domain shifts are substantial, the gradient updates for the global model during each round deviate significantly from the invariant gradient direction. This hinders straightforward convergence to the global optimum due to its simplistic optimization approach.}
    \vspace{0.5cm}
    \label{fig:demo}
\end{figure}

Recently, gradient-based DG has emerged as a promising approach to enhance generalization across domains. {This approach is \emph{orthogonal} to the previously mentioned methods in nature, allowing for their \emph{combined integration to facilitate additional performance improvements}.}
Particularly, gradient-based DG's target is discovering an invariant gradient direction across source domains. 
Fish~\cite{2022-DG-Fish} introduces a gradient inner product (GIP) to ensure consistent gradient trajectories among domains, whereas Fishr \cite{2022-DG-Fishr} minimizes the Euclidean distance between domains' gradient trajectories. 
These methods increase the likelihood that the gradient direction remains consistent even on unseen target domains. 
In this paper, we perform a theoretical and empirical analysis to show that there exists the potential degradation in the learning progress of these methods due to \emph{gradient fluctuation} (see the concept in Fig.~\ref{fig:demo}). 
We show that gradient fluctuation is caused by updating each domain sequentially in Reptile \cite{2018-MeL-Reptile} based Fish \cite[Alg. 1]{2022-DG-Fish}.
Additionally, employing Hessian approximations, as in Fish \cite[Alg. 2]{2022-DG-Fish}, incurs substantial computational overhead.
3
To address these challenges, we introduce Pareto Optimality Gradient Matching (POGM), a simple but effective gradient-based domain generalization method.
\textbf{First}, to mitigate gradient fluctuation phenomenon, we formulate our gradient matching by summing over pairs of gradient inner product (GIP), and we confine the search space for the GIP solution to a $\kappa$-hypersphere centered around the ERM gradient trajectory to reduce the effort in finding optimal solutions.
\textbf{Second}, to solve the summation of GIP of gradients over $K$ domains, which has a complexity of $\mathcal{O}(K\times(K-1)/2)$, we first utilize the Pareto front to transform the task of minimizing all GIP pairs into focusing solely on the worst-case scenario. Then, we introduce a closed-form relaxation method for inter-domain gradient matching. As a result, the complexity of our POGM can be reduced to $\mathcal{O}(2\times K)$.
\textbf{Third}, to circumvent the computational overhead associated with Hessian approximations, we leverage meta-learning and consider gradient matching as a separate process. Hence, our method can learn a set of coefficients for combining domain-specific gradients with scaled weights. As a consequence, POGM can approximate weighted aggregated domain-specific gradient updates without the need for second-order derivatives.
Our experiments show that POGM achieves state-of-the-art performance across datasets from the recent DG benchmark, DomainBed \cite{2021-DG-DomainBed}. The robust performance of our method across diverse datasets underscores its broad applicability to different applications and sub-genres of DG tasks.

\section{Related Works}
Several approaches for DG have been explored, which can be broadly classified into two categories: finding domain-invariant representation and representation mixing. From the perspective of the first category, \cite{2021-DG-CausalMatching} introduces a regularization-based framework to generate invariant representations by minimizing the empirical distance between encoded representations from different domains. \cite{2022-DG-KLGuided,2021-DG-IRL-DDT} minimize the empirical distance between the source and target data distribution. \cite{2021-DG-IRL-DDT} assumed that the target dataset is not accessible and therefore proposed to generate data on the target dataset using a generative adversarial network (GAN). However, this approach is applicable within limited constraints, where the target dataset is required to be accessible to the GAN model. \cite{2018-DG-CIAN,2020-DG-EntropyReg,2021-DG-DomainSpecificFeatures} propose GAN-based frameworks for learning invariant representations. These works highlight the redundancy in classifier networks, leading to the introduction of a discriminator network to enhance the extraction of meaningful information.
Additionally, \cite{2021-DG-SelfReg} suggests leveraging self-supervised learning to generate domain-invariant representations. \cite{2022-DG-CIRL} applies Barlow Twins to generate causal invariant representations, assumed to be domain-invariant. According to the second category, \cite{2021-DG-DAML,2022-DG-CMixup,2021-DG-MixStyle,2020-DG-Mixup} leverages a mixing strategy to inter-domain representations to improve the DG. \cite{2021-DG-FeatureAugmentation} proposes a simple data augmentation approach by perturbing the latent features with white Gaussian noise. 

SWAD \cite{2021-DG-SWAD} proposes a different approach for DG by leveraging stochastic weight averaging to smooth the loss landscape, thus improving the generalization.

Recently, \cite{2022-DG-Fish} pioneers gradient-based DG, introducing Fish to discover invariant gradient trajectories across domains for enhanced model consistency amidst domain shifts, thereby improving generalization to unseen datasets. Additionally, \cite{2022-DG-Fishr} introduces Fishr to enhance gradient learning by incorporating gradient variant regularization into the loss function, capturing both the first and second moments of the gradient distribution, thereby leveraging richer information for gradient learning. 
Our work is also a gradient-based DG method, in which we aim to address the gradient limitations in Fish and Fishr by exploiting Pareto optimality for gradient matching. 
\section{Problem Settings}\label{sec:problem-settings}
Let $\mathcal{X}$ and $\mathcal{Y}$ be the feature and label spaces, respectively. 
There are $K$ source domains $\mathcal{K} = \{D_i\}^{K}_{i=1}$ and $L$ target domains $\{D_i\}^{K+L}_{i=K+1}$. 
The goal is to generalize the model learned using data samples of the source domains to unseen target domains. 
Herein, we denote the joint distribution of domain $i$ by $P_i(X, Y)$ ($X, Y\sim \mathcal{X},\mathcal{Y}$). 
During training, there are $K$ datasets $\{S_i\}^{K}_{i=1}$ available, where $S_i=\{(x^{(i)}_j,y^{(i)}_j)\}^{N_i}_{j=1}$, $N_i$ is the number of samples of $S_i$ that are sampled from the $i^{\textrm{th}}$ domain. 
At test time, we evaluate the generalization capabilities of the learned model on $L$ datasets sampled from the $L$ target domains, respectively. 
This work focuses on DG for image classification where the label space $\mathcal{Y}$ contains $C$ discrete labels $\{1,2,\ldots, C\}$.

To facilitate our analysis, we model a gradient-based domain generalization (GBDG) algorithm that aims to learn an invariant gradient trajectory via the generalized loss $\mathcal{L}_\textrm{GBDG} = \mathcal{L}_\textrm{ERM} + \lambda\mathcal{L}_\textrm{IG},$ where $\mathcal{L}_\textrm{GBDG}$ is denoted as GBDG loss,
$\mathcal{L}_\textrm{ERM}$ is the empirical risk minimization (ERM) \cite{1998-DG-ERM} loss function.
$\mathcal{L}_\textrm{IG}$ is the invariant gradient regularizer, which can represent both Fish~\cite{2022-DG-Fish} or Fishr~\cite{2022-DG-Fishr}, and can be defined as 
$\mathcal{L}_\textrm{IG} = \sum_{i,j\in \mathcal{K}}\mathcal{L}_\textrm{dist}(\nabla\mathcal{L}_i(\theta), \nabla\mathcal{L}_j(\theta)),$
where $\mathcal{L}_\textrm{dist}$ denotes the empirical distance between two gradient vectors of the model parameters $\theta$ trained from domains $i$ and $j$, respectively. We denote $\mathcal{L}_i$ as the loss according to the training on a domain $i\in \mathcal{K}$. $\mathcal{L}_\textrm{dist}$ is computed via GIP and by mean-square error (MSE) in Fish and Fishr, respectively. To understand Fish and Fishr, we conduct both theoretical analysis and empirical experiments on three baselines, including Fish, Fishr, and the conventional ERM. The detailed setup of the experiments is demonstrated in Appendix~\ref {app:analysis-setup}.

\begin{figure}[!h]
    \begin{subfigure}[b]{0.25\textwidth}
        \includegraphics[width=\textwidth]{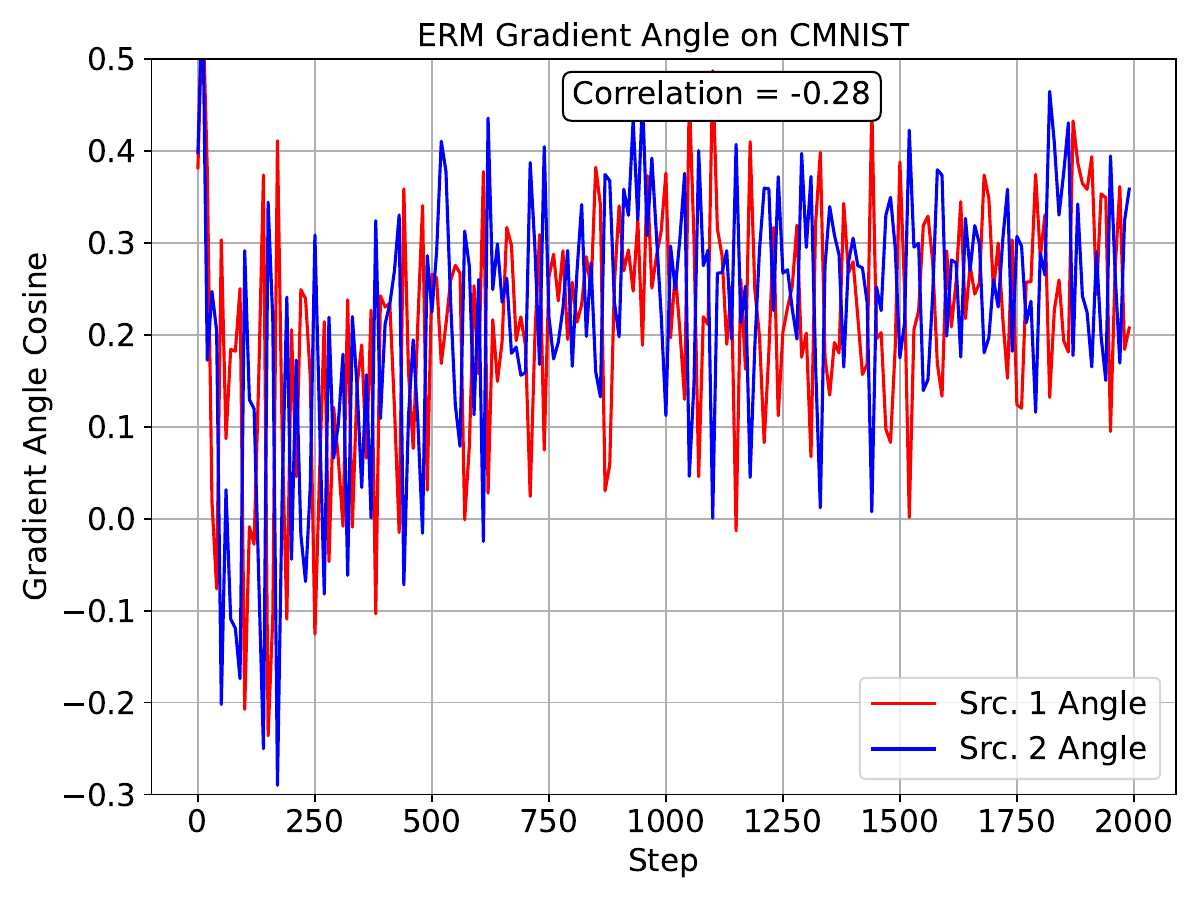}
        \caption{}
        \vspace{0.5cm}
        \label{fig:cmnist-erm-angle}
    \end{subfigure}%
    \begin{subfigure}[b]{0.25\textwidth}
        \includegraphics[width=\textwidth]{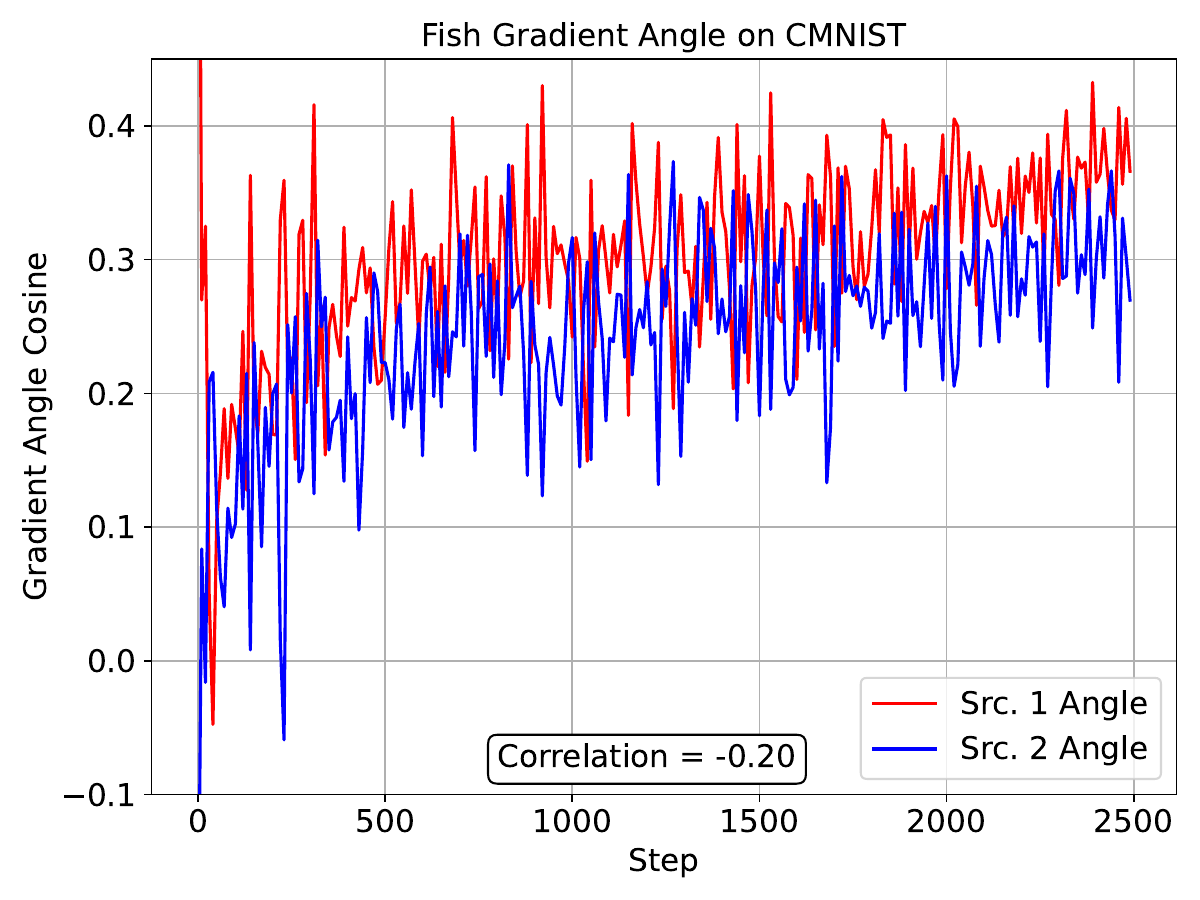}
        \caption{}  
        \vspace{0.5cm}
        \label{fig:cmnist-fish-angle}
    \end{subfigure}
    \begin{subfigure}[b]{0.25\textwidth}
        \includegraphics[width=\textwidth]{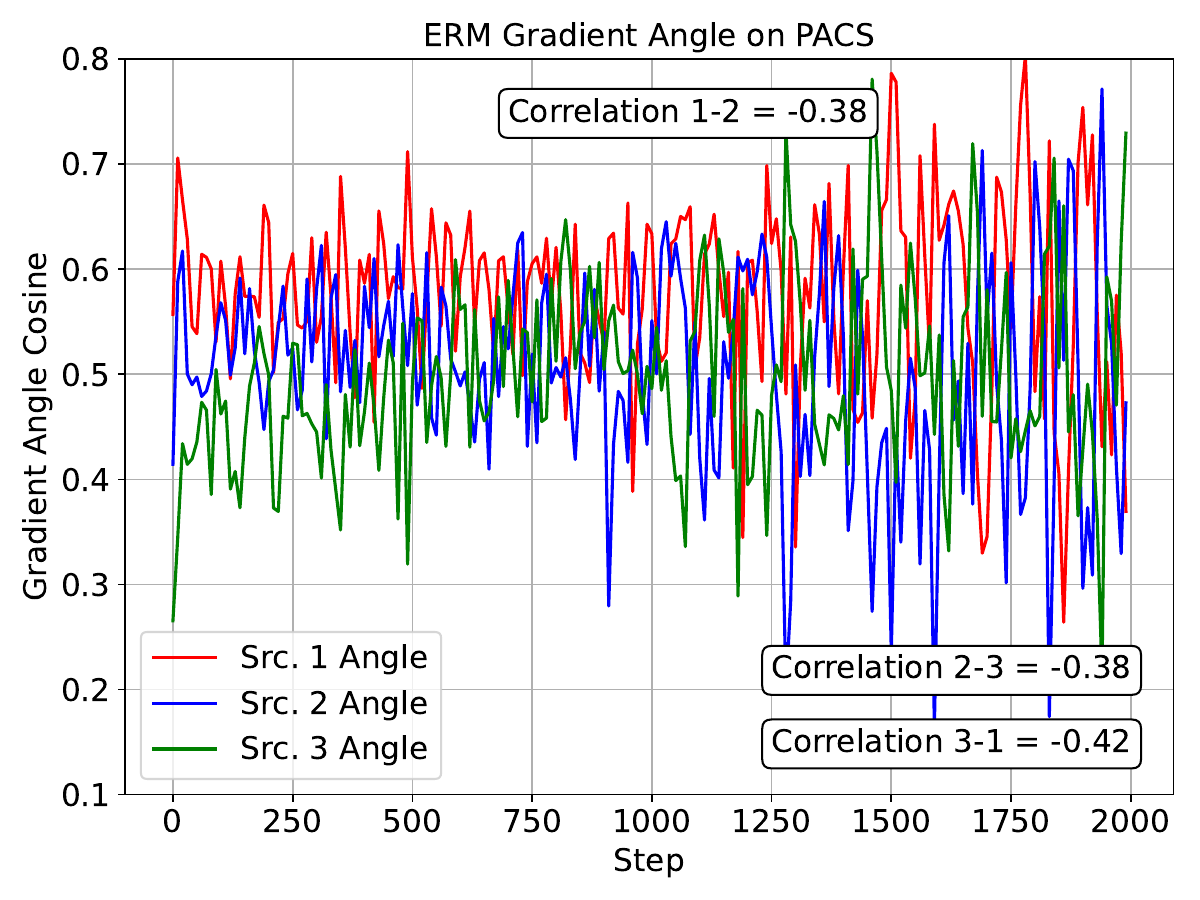}
        \caption{}
        \label{fig:pacs-erm-angle}
    \end{subfigure}%
    \begin{subfigure}[b]{0.25\textwidth}
        \includegraphics[width=\textwidth]{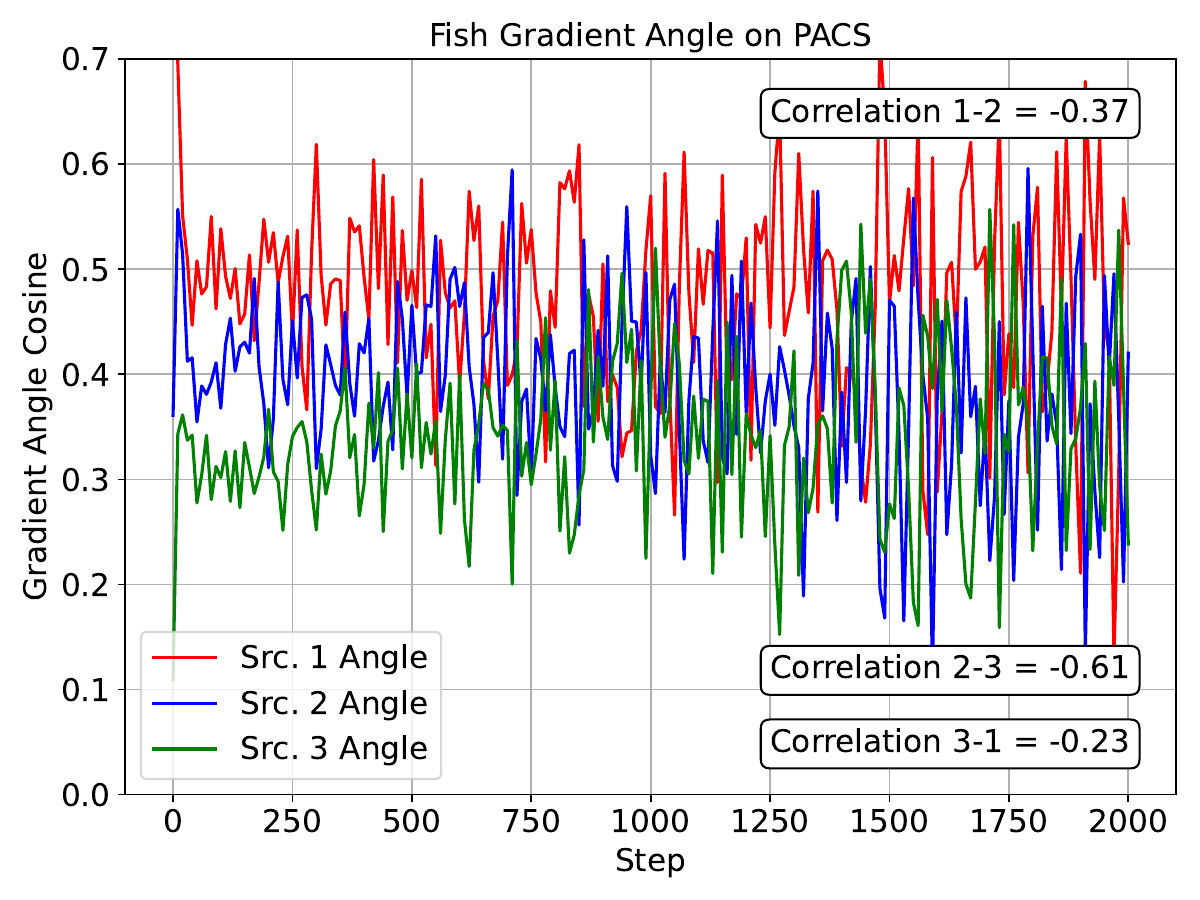}
        \caption{}  
        \label{fig:pacs-fish-angle}
    \end{subfigure}
    \vspace{0.3cm}
    \caption{The gradient direction angles of ERM and Fish in CMNIST, PACS. Although it is claimed that the Fish gradient is invariant to domain-specific gradients, the angles between the Fish gradient and domain-specific gradients tend to fluctuate considerably. This leads to a lower correlation among domain-specific gradients compared to the ERM gradient.}
    \vspace{0.5cm}
    \label{fig:fish-erm-angle}
\end{figure}


We show that although the conventional Fish \cite[Alg.~2]{2022-DG-Fish} aims to reduce the gradient angles via GIP, the Reptile-based Fish \cite[Alg.~1]{2022-DG-Fish} yields significantly fluctuating gradient directions (i.e., lower correlation among angles' cosines). 
This can be observed in both the CMNIST and PACS datasets (see Fig.~\ref{fig:fish-erm-angle}), where Fish shows no significant improvement compared with the conventional ERM. 

\section{POGM: Gradient Matching via Pareto Optimality}

\subsection{Gradient Inner Product with Generalized Constraints}
Our method aims to simultaneously alleviate the issues of gradient fluctuation (Section~\ref{sec:problem-settings}). We adopt GIP as proposed from Fish \cite{2022-DG-Fish} but aim to restrict the searching space for our GIP problem within a $\kappa$-hypersphere, which has the center determined by the ERM trajectory. Specifically, we propose the GIP with generalized constraints (GIP-C) as follows:
\begin{align}
    \mathcal{L}_\textrm{GIP-C} 
    & = \sum_{i\in\mathcal{K}}\sum^{i\neq j}_{j\in\mathcal{K}}\nabla\mathcal{L}_i(\theta)\cdot\nabla\mathcal{L}_j(\theta) \nonumber \\
    &\quad - \gamma\Big(\Vert\nabla\mathcal{L}_\textrm{GIP-C} - \nabla\mathcal{L}_\textrm{ERM}\Vert^2 - \kappa\Vert\nabla\mathcal{L}_\textrm{ERM}\Vert^2\Big),
\label{eq:GIP-C-N}
\end{align}

where $\mathcal{L}_\textrm{ERM}$ is the ERM loss defined as $\mathcal{L}_\textrm{ERM} = \frac{1}{K}\sum^{K}_{i=1}\mathcal{L}_i(\theta)$. We utilize ERM as a standard gradient trajectory as ERM is simple, straightforward, and demonstrates good results in DG. The advantage of adding this constraint is twofold: \textbf{1)} The learned invariant gradient $\nabla\mathcal{L}_\textrm{GIP-C}$ is not biased to one set of domains, which ensures the generalization of our algorithm. \textbf{2)} The learned invariant gradient $\nabla\mathcal{L}_\textrm{GIP-C}$ is either backtracked towards the previous checkpoint or biased towards specific domains with a value larger than $\kappa\Vert\nabla\mathcal{L}_\textrm{ERM}\Vert$.

However, the incorporation of loss as depicted in \eqref{eq:GIP-C-N} necessitates a second-order derivative approximation and becomes NP-hard. Additionally, the algorithm has a complexity of $\mathcal{O}(K\times(K-1)/2)$ as it sums over all pairs of gradients in $K$ domains. Consequently, the learning is hampered by both the computational overhead and the challenge of identifying the optimal solution. 

In the next section, our target advances two objectives. First, to alleviate the NP-hard problem, we propose a relaxation for the optimization problem to reduce the computational complexity. Second, to mitigate computational overhead caused by approximating the second-order derivative, we utilize meta-learning \cite{2017-MeL-MAML} to establish a unified learning framework for our DG process.
\begin{algorithm}
\caption{Domain Generalization via Pareto Optimality Gradient Matching}\label{alg:POGM}
    \SetKwInOut{KwIn}{Input}
    \SetKwInOut{KwOut}{Output}
    \KwIn{ Number of training domain $K$, initial model parameter $\theta^{(0)}$, learning rate $\alpha$.}
    \KwOut{ Model parameters $\theta$}
    \For{\textit{each round} $r=0,\ldots, R$}{
            \For{domain $i\in \mathcal{K}$}{
                Append the meta model to the domain-wise model $\theta^{(r,0)}_i\leftarrow \theta^{(r)}$. \\
                \For{local epoch $e \in E$}{
                    Sample mini-batch $\zeta$ from local data $S_i$ \\
                    Update domain-wise model $\theta^{(r,e+1)}_i = \theta^{(r,e)}_i - \eta \nabla \mathcal{L}_i(\theta^{(r,e)}_i, \zeta)$. \\
                }
            }
        Apply \textbf{Meta Update}
        }
    \SetKwProg{Ser}{Meta Update}{}{}
    \Ser{}{
    Calculate $h^{(r)}_{i}$ from $\theta^{(r,E)}_i$, $\theta^{(r)}_i$ according to $h^{(r)}_{i} = \theta^{(r,E)}_i-\theta^{(r)}_i$. \\
    Calculate $h^{(r)}_{\textrm{ERM}} = \frac{1}{K}\sum^{K}_{i=1} h^{(r)}_{i}$ as the average gradient update. \\
    At the $r^{th}$ optimization, $\phi = \kappa^2 \Vert h^{(r)}_{\textrm{ERM}}\Vert^2$. \\
    Find the optimal $\widetilde{\pi}$ set by solving $\widetilde{\pi} = \arg\min_{\pi} h^{(r)}_\pi\cdot h^{(r)}_{\textrm{ERM}} + \sqrt{\kappa}\Vert h^{(r)}_{\textrm{ERM}}\Vert\Vert h^{(r)}_{\pi}\Vert$, where $h^{(r)}_{\pi} = \sum^{K}_{i=1} \pi_i h^{(r)}_{i}$, $\pi_i \in [0,1],~\sum^{K}_{i=1}\pi_i = 1,~\forall i\in \mathcal{K}$ \\
    Update 
    \begin{align}
        \theta^{(r+1)} = \theta^{(r)} - \alpha h_\textrm{GIP-C}, \notag
    \end{align}                             where $h_\textrm{GIP-C}$ is defined via Theorem~\ref{theorem:surrogate-IDGM}.
    }
\end{algorithm}
\subsection{Relaxation of Inter-domain Gradient Matching}\label{sec:relaxation}
By finding $\theta_\textrm{GIP-C}$ using \eqref{eq:GIP-C-N}, we can find $\theta_\textrm{GIP-C}$ that achieve the angles between gradients induced by $\theta_\textrm{GIP-C}$ to the gradients on source domains $D_k, \forall i\in \mathcal{K}$ are maximized. For instance, 
\begin{align}
    \theta_\textrm{GIP-C} 
    & = \arg\max_{\theta}\sum^{}_{i\in\mathcal{K}}
    \nabla\mathcal{L}_i(\theta)\cdot\nabla\mathcal{L}_\textrm{GIP-C} \nonumber \\
    & \quad - \gamma\Big(\Vert\nabla\mathcal{L}_\textrm{GIP-C} - \nabla\mathcal{L}_\textrm{ERM}\Vert^2 - \kappa\Vert\nabla\mathcal{L}_\textrm{ERM}\Vert^2\Big).
\label{eq:GIP-C-2}
\end{align}
Herein, \eqref{eq:GIP-C-2} is a multi-objective optimization problem. In general, no single solution can optimize all objectives at the same time. To overcome this, we determine the Pareto front that provides a trade-off among the different objectives. We consider the following definitions \cite{1999-OPT-Pareto}: 
\begin{definition}[Pareto dominance]
    Let $\theta^{a}, \theta^{b} \in R^{m}$ be two points, $\theta^{a}$ is said to be dominated $\theta^{b}$ ($\theta^{a}\succ\theta^{b}$) if and only if $\mathcal{L}_i(\theta^{a})\leq\mathcal{L}_i(\theta^{b}), \forall i\in\mathcal{K}$ and $\mathcal{L}_j(\theta^{a})<\mathcal{L}_j(\theta^{b}), \exists j\in\mathcal{K}$.
\label{def:pareto-dominance}
\end{definition}
\begin{definition}[Pareto optimality]
    $\theta^{*}$ is a Pareto optimal point and $\mathcal{L}(\theta^{*})$ is a Pareto optimal objective vector if it does not exist $\hat{\theta}\in R^{m}$ such that $\hat{\theta}\prec\theta^{*}$. The set of all Pareto optimal points is called the Pareto set. The image of the Pareto set in the loss space is called the Pareto front.
\label{def:pareto-optimality}
\end{definition}
To leverage Definition~\ref{def:pareto-optimality}, we present the following lemma:
\begin{lemma}
    The average cosine similarity between the given gradient vector $\nabla\mathcal{L}_\textrm{GIP-C}$ and the domain-specific gradient is lower-bounded by the worst-case cosine similarity as follows:
    \begin{align}
        \frac{1}{K}\sum^{}_{i\in \mathcal{K}}\nabla\mathcal{L}_i(\theta)\cdot\nabla\mathcal{L}_\textrm{GIP-C} \geq \min_{i\in\mathcal{K}}\nabla\mathcal{L}_i(\theta)\cdot\nabla\mathcal{L}_\textrm{GIP-C}.\notag
    \end{align}
\label{lemma:worst-case-grad}
\end{lemma}
Lemma~\ref{lemma:worst-case-grad} allows the realization that the maximization of our multi-objective function can be reduced to maximizing the worst-case scenario. The approach leads us to attain the optimal Pareto front. Hence, the following lemma follows:
\begin{lemma}
    Given $\theta^{*}$ as an Pareto optimal solution of $\theta$, we have $\theta^{*}$, which is also the solution of
    \begin{align}
        &\max_{\theta}\min_{i\in\mathcal{K}}
        \Bigg[ \nabla\mathcal{L}_i(\theta)\cdot\nabla\mathcal{L}_\textrm{GIP-C}(\theta) \\
        & \quad - \gamma\Big(\Vert\nabla\mathcal{L}_\textrm{GIP-C}(\theta) - \nabla\mathcal{L}_\textrm{ERM}(\theta)\Vert^2 - \kappa\Vert\nabla\mathcal{L}_\textrm{ERM}(\theta)\Vert^2\Big)\Bigg] \nonumber
    \end{align}
\label{lemma:GIP-C-Pareto} 
\end{lemma}
Lemma~\ref{lemma:GIP-C-Pareto} represents the optimization over the gradient. To simplify implementation and reduce computational complexity, we seek optimal gradients along gradient trajectories. The approach enables us to circumvent noise introduced by mini-batch gradients, ensuring both optimization accuracy and stability. Specifically, we define the gradient trajectory of domain $i$ and define the ERM gradient trajectory as
\begin{align}
    & h^{(r)}_i = \theta^{(r+1)}_i - \theta^{(r)} = \sum^{E}_{e=1} \nabla\mathcal{L}_i (\theta^{(r,e)}),
    \nonumber \\
    & h^{(r)}_\textrm{ERM} = \theta^{(r+1)}_\textrm{ERM} - \theta^{(r)}_\textrm{ERM} = \frac{1}{K}\sum^{K}_{i=1}\sum^{E}_{e=1}\nabla\mathcal{L}_i (\theta^{(r,e)}). 
\label{eq:grad-trajectory}
\end{align}
From Lemma~\ref{lemma:GIP-C-Pareto} on Pareto optimal solution, we derive the theorem for an invariant gradient solution as follows:
\begin{theorem}[Invariant Gradient Solution]
    Given the Pareto condition as mentioned in Lemma~\ref{lemma:GIP-C-Pareto}, $\pi = \{\pi^{(r)}_{1},\ldots,\pi^{(r)}_{K}\}$ are the set of $K$ learnable scaling parameters, which coordinate the domain-wise gradient trajectory $h^{(r)}_{i},~\forall i\in\mathcal{K}$ at each training iteration. The invariant gradient $h_\textrm{GIP-C}$ is characterized by: 
    \begin{align}
        & h^{(r)}_\textrm{GIP-C} = h^{(r)}_{\textrm{ERM}} + \frac{\kappa\Vert h^{(r)}_{\textrm{ERM}}\Vert}{\Vert h^{(r)}_{\widetilde{\pi}}\Vert}h^{(r)}_{\widetilde{\pi}} \nonumber\\
        & \quad \textrm{s.t.} 
        \quad 
        \widetilde{\pi} = \arg\min_{\pi} h^{(r)}_\pi\cdot h^{(r)}_{\textrm{ERM}} + \sqrt{\kappa}\Vert h^{(r)}_{\textrm{ERM}}\Vert\Vert h^{(r)}_{\pi}\Vert,
    \end{align}
    where $h^{(r)}_{\pi} = \sum^{K}_{i=1}\pi^{(r)}_i h^{(r)}_i,~\sum^{K}_{i=1}\pi^{(r)}_i = 1$. We denote $\widetilde{\pi}$ as the optimal parameter set at round $r$.
\label{theorem:surrogate-IDGM}
\end{theorem}
\begin{remark}
    The computation of the loss function using Theorem~\ref{theorem:surrogate-IDGM} reduces to $\mathcal{O}(2\times K)$ as we only need to compute the GIP between two aggregated gradients once.
\end{remark}
From Theorem~\ref{theorem:surrogate-IDGM}, GIP-C appears to have a close relationship with ERM. For instance,
\begin{corollary}
    When the radius of the $\kappa$-hypersphere reduces to $0$, the GIP-C is reduced to ERM. For instance, $\lim_{\kappa\rightarrow 0} h_{\textrm{GIP-C}} = h^{(r)}_{\textrm{ERM}}.$
\end{corollary}
Furthermore, leveraging Pareto optimality \cite{1999-OPT-Pareto}, we derive a corollary as follows:
\begin{corollary}
    The optimal GIP-C solution is always better than that of the optimal ERM solution. For instance, $\mathcal{L}({\theta^{*}_{\textrm{GIP-C}}}) > \mathcal{L}({\theta^{*}_{\textrm{ERM}}})$.
\end{corollary}
Hence, GIP-C consistently exhibits superior performance compared with ERM, contributing as one of the most effective baselines to date.
To determine the invariant gradient trajectory, we propose a Meta-learning-based approach \cite{2017-MeL-MAML} for our DG. Firstly, the agent aims to train its model parameter using domain-wise data via the optimization problem $\theta^{(r)}_i = \arg\min_{\theta} \mathcal{L}(\theta^{(r)}, S_i)$ at the local stage. Therefore, the domain-wise gradient trajectory can be computed via \eqref{eq:grad-trajectory}. At the meta update stage, the agent leverages the domain-wise gradients to approximate the invariant gradient trajectory $h_\textrm{GIP-C}$ using Theorem~\ref{theorem:surrogate-IDGM}. Thereafter, the model is updated using updating function $\theta^{(r+1)} = \theta^{(r)} - \alpha h^{(r)}_\textrm{GIP-C}$. The detailed algorithm is presented in Alg.~\ref{alg:POGM}.

\subsection{Theoretical Analysis}\label{sec:theoretical-analysis}
We consider two theoretical analyses for our proposed method in accordance to gradient invariant and DG properties. For instance,
\begin{theorem}[Gradient Invariant Properties]
    Given $U_i = \nabla\mathcal{L}_i(\theta)\cdot\nabla\mathcal{L}_\textrm{GIP-C}(\theta)$ as the utility function of the Pareto Optimality problem. At each round $r$ where the Pareto is applied on different domains $i$, we have the GIP variance reduced accordingly to the number of learned epochs applied. For instance, $\textrm{Var}\Big(U_i(\theta^{(r+1,e)})\Big)
    \leq {\textrm{Var}\Big(U_i(\theta^{(r+1,e)})\Big)}\Big/{E^{*}\Big(\frac{\eta^2 L}{2} - \eta\Big)}.$
\label{theorem:invariant-property}
\end{theorem}
\begin{remark}
    The number of epochs $E$ is related to the progress in gradient magnitude each round. Thus, GIP variance reduction also relates to the gradient variance $\Big\Vert \nabla U_k(\theta^{(r,e)})\Big\Vert^2$. As the gradient variance decreases, the gradient step becomes smaller, and thus the gradient invariant analysis becomes more stable.
\end{remark}
\begin{lemma}[Divergence of domains in the sources' convex hull \newline\cite{2021-DG-G2DM}]
    Let $d_\mathcal{V}[\mathcal{D}_{i}, \mathcal{D}_{j}] \leq\epsilon$, $\forall i, j\in [K]$. $\sphericalangle_{K}$ is the convex hull formed by source domains. The following inequality holds for the $\mathcal{V}$-divergence between any pair of domains $\mathcal{D}^{'}, \mathcal{D}^{''} \in \sphericalangle_{K}$: $d_\mathcal{V}[\mathcal{D}^{'}, \mathcal{D}^{''}]\leq\epsilon$.
\label{lemma:divergence-convex-hull}
\end{lemma}
From these two lemmas, we have the risk bounds on DG:
\begin{theorem}[Optimal generalized risk bounds]
    If a target domain lies beneath the convex hull formed by $K$ source domains, then we can achieve the optimal generalized risk on target domains when the following holds:

\begin{align}
    \LL^{*}_\textrm{L} 
    = \LL^{*}_\textrm{K} 
    & + 
    \frac{M}{2K} \sqrt{\sum^{K}_{i=1}\sum^{K}_{j=1} D_\textrm{KL}\Big[p_i(y\vert x,\theta^{*}) \,\Big\Vert\, p_j(y\vert x,\theta^{*})\Big]} \notag \\
    & + \frac{M}{2K} \sqrt{\sum^{K}_{i=1}\sum^{K}_{j=1} D_\textrm{KL}\Big[p_i(x\vert\theta^{*}) \,\Big\Vert\, p_j(x\vert\theta^{*})\Big]} \notag \\
    &\textrm{s.t.}~~~ \theta^{*} = \underset{\theta}{\arg\max} \sum^{i\neq j}_{i,j\in\mathcal{K}}\nabla\mathcal{L}_i(\theta) \cdot \nabla\mathcal{L}_j(\theta),
\end{align}

where $\LL^{*}_\textrm{K}, \LL^{*}_\textrm{L}$ are the source and target risks, respectively.
\label{theorem:domain-convex-hull}
\end{theorem}
Theorem~\ref{theorem:domain-convex-hull} demonstrates that by maximizing the GIP among domain gradients during each meta update round, we can effectively reduce the generalization gap between the source and target domains.
\begin{remark}
    The term $D_\textrm{KL}\Big[p_i(x\vert\theta^{*})\Vert p_j(x\vert\theta^{*})\Big]$ represents the distribution of the dataset between two domains $i,j$, and is independent of $\theta^{*}$. Therefore, we can consider this term irreducible.
\end{remark}
\begin{remark}
    The term $D_\textrm{KL}\Big[p_i(y\vert x,\theta^{*})\Vert p_j(y\vert x,\theta^{*})\Big]$ refers to the gap between the two hypotheses $p_i(y\vert x,\theta^{*}), p_j(y\vert x,\theta^{*})$. This term is optimizable, and in our work, we optimize it by finding the model $\theta^{*}$ that maximizes the similarity among domain-specific gradients.
\end{remark}
\section{Experimental Evaluation}
\subsection{Results on DomainBed Benchmark}\label{sec:overall-benchmarks}
\begin{table*}[!ht]
\centering
\caption{DomainBed benchmark. We format \textbf{first}, \underline{second}, and \textcolor{gray}{worse than ERM results}.}
\label{tab:domainbed-benchmark}
\adjustbox{max width=\textwidth}{%
\begin{tabular}{l|cccccccc|lll}
\toprule
\multirow{2}{*}{\textbf{Algorithm}} & \multicolumn{7}{c}{Accuracy($\uparrow$)}                                                                                                             &              & \multicolumn{3}{c}{Ranking($\downarrow$)}                                                                                                        \\
                                    & \textbf{CMNIST} & \textbf{RMNIST} & \textbf{VLCS}  & \textbf{PACS}  & \textbf{OfficeHome} & \textbf{TerraInc.} & \textbf{DomainNet} & \textbf{Avg} & \textbf{\begin{tabular}[c]{@{}l@{}}Arith.\\ mean\end{tabular}} & \textbf{\begin{tabular}[c]{@{}l@{}}Geom.\\ mean\end{tabular}} & \textbf{Median} \\ 
\midrule
ERM                                 & 57.8 $\pm$ 0.2        & 97.8 $\pm$ 0.1        & 77.6 $\pm$ 0.3 & 86.7 $\pm$ 0.3 & 66.4 $\pm$ 0.5      & 53.0 $\pm$ 0.3          & 41.3 $\pm$ 0.1     & 68.7         & 9.00&	8.0&	10\\
RSC \cite{2020-DG-RSC} (ECCV, 2020) & 58.5 $\pm$ 0.5        &  \textcolor{gray}{97.6} $\pm$ 0.1        & 77.8 $\pm$ 0.6 & \textcolor{gray}{86.2} $\pm$ 0.5 & 66.5 $\pm$ 0.6      &  \textcolor{gray}{52.1} $\pm$ 0.2          &  \textcolor{gray}{38.9} $\pm$ 0.6     & \textcolor{gray}{68.2}         &  \textcolor{gray}{10.6}                                                            &  \textcolor{gray}{10.2}                                                           &  \textcolor{gray}{11}               \\
MTL \cite{2021-DG-MTL} (JMLR, 2021) &  \textcolor{gray}{57.6} $\pm$ 0.3        & 97.9 $\pm$ 0.1        & 77.7 $\pm$ 0.5 & 86.7 $\pm$ 0.2 & 66.5 $\pm$ 0.4      &  \textcolor{gray}{52.2} $\pm$ 0.4          &  \textcolor{gray}{40.8} $\pm$ 0.1     & \textcolor{gray}{68.5}         & {8.4}&	{7.7}	&{7}
 \\
SagNet \cite{2021-DG-SagNet} (CVPR, 2021) & 58.2 $\pm$ 0.3        & 97.9 $\pm$ 0.0        & 77.6 $\pm$ 0.1 & \textcolor{gray}{86.4} $\pm$ 0.4 & 67.5 $\pm$ 0.2      &  \textcolor{gray}{52.5} $\pm$ 0.4          &  \textcolor{gray}{40.8} $\pm$ 0.2     & 68.7         & {8.0}	& {7.3}&	{7}             \\
ARM \cite{2021-DG-ARM} (NIPS, 2021) & 63.2 $\pm$ 0.7        & \underline{98.1} $\pm$ 0.1        & 77.8 $\pm$ 0.3 & \textcolor{gray}{85.8} $\pm$ 0.2 &  \textcolor{gray}{64.8} $\pm$ 0.4      &  \textcolor{gray}{51.2} $\pm$ 0.5          &  \textcolor{gray}{36.0} $\pm$ 0.2     & \textcolor{gray}{68.1}         &  \textcolor{gray}{10.0}                                                            & \textcolor{gray}{8.3}  &  {10}              \\
VREx \cite{2021-DG-VREx} (ICML, 2021) & 67.0 $\pm$ 1.3        & 97.9 $\pm$ 0.1        & 78.1 $\pm$ 0.2 & \underline{87.2} $\pm$ 0.6 &  \textcolor{gray}{65.7} $\pm$ 0.3      &  \textcolor{gray}{51.4} $\pm$ 0.5          &  \textcolor{gray}{30.1} $\pm$ 3.7     & \textcolor{gray}{68.2}         & {8.1}&	{6.1}&	{7} \\
AND-mask (ICLR, 2021) & 58.6 $\pm$ 0.4        &  \textcolor{gray}{97.5} $\pm$ 0.0        &  \textcolor{gray}{76.4} $\pm$ 0.4 & \textcolor{gray}{86.4} $\pm$ 0.4 &  \textcolor{gray}{66.1} $\pm$ 0.2      &  \textcolor{gray}{49.8} $\pm$ 0.4          &  \textcolor{gray}{37.9} $\pm$ 0.6     & \textcolor{gray}{67.5}         &  \textcolor{gray}{13.0}                                                           &  \textcolor{gray}{12.7}                                                          &  \textcolor{gray}{13}              \\
SAND-mask \cite{2020-DG-SandMask} (ICML, 2021) & 62.3 $\pm$ 1.0        &  \textcolor{gray}{97.4} $\pm$ 0.1        &  \textcolor{gray}{76.2} $\pm$ 0.5 & \textcolor{gray}{85.9} $\pm$ 0.4 &  \textcolor{gray}{65.9} $\pm$ 0.5      & \textcolor{gray}{50.2} $\pm$ 0.1          &  \textcolor{gray}{32.2} $\pm$ 0.6     & \textcolor{gray}{67.2}         &  \textcolor{gray}{14.0}                                                           &  \textcolor{gray}{13.3}                                                          &  \textcolor{gray}{13}              \\
{EQRM \cite{2022-DG-EQRM} (NIPS, 2022)}          & \textcolor{gray}{53.4} $\pm$ 1.7        & 98.0 $\pm$ 0.0        & {77.8} $\pm$ 0.6 & \textcolor{gray}{86.5} $\pm$ 0.2 &  {67.5} $\pm$ 0.1      &  \textcolor{gray}{47.8} $\pm$ 0.6          & \textcolor{gray}{41.0} $\pm$ 0.3     & {67.4}         &  {8.99} & 7.4&  {7}            \\
{CIRL \cite{2022-DG-CIRL} (CVPR, 2022)}          & 62.1 $\pm$ 1.3        & 97.7 $\pm$ 0.2        & {78.6} $\pm$ 0.6 & \textcolor{gray}{86.3} $\pm$ 0.4 &  {67.1} $\pm$ 0.3      &  \textcolor{gray}{52.1} $\pm$ 0.2          & \textcolor{gray}{39.7} $\pm$ 0.4     & {69.1}         &  {8.7} & \textcolor{gray}{8.1}&  {7}            \\
{Mixstyle \cite{2021-DG-MixStyle} (ICLR, 2022)}          & \textcolor{gray}{54.4} $\pm$ 1.4        & 97.9 $\pm$ 0.0        & 77.9 $\pm$ 0.1 & \textcolor{gray}{85.2} $\pm$ 0.1 &  \textcolor{gray}{60.4} $\pm$ 0.5      &  \textcolor{gray}{44.0} $\pm$ 0.6          & \textcolor{gray}{34.0} $\pm$ 0.3     & \textcolor{gray}{64.8}         &  \textcolor{gray}{14.0} & \textcolor{gray}{12.8}&  \textcolor{gray}{17}            \\
Fish (ICLR, 2022) & 61.8 $\pm$ 0.8        & 97.9 $\pm$ 0.1        & 77.8 $\pm$ 0.6 & \textcolor{gray}{85.8} $\pm$ 0.6 &  \textcolor{gray}{66.0} $\pm$ 2.9      & \textcolor{gray}{50.8} $\pm$ 0.4          & \textbf{43.4} $\pm$ 0.3     & 69.1         & 8.6&	6.8&	9\\
Fishr (ICML, 2022) & \textbf{68.8} $\pm$ 1.4        & 97.8 $\pm$ 0.1        & 78.2 $\pm$ 0.2 & 86.9 $\pm$ 0.2 & {68.2} $\pm$ 0.2      & {53.6} $\pm$ 0.4          & 41.8 $\pm$ 0.2     & \underline{70.8}         & \underline{4.6}	&\underline{3.6}&	\underline{3}\\
{MADG (NIPS, 2023)}          & 60.4 $\pm$ 0.8        & 97.9 $\pm$ 0.0        & 78.7 $\pm$ 0.2 & \textcolor{gray}{86.5} $\pm$ 0.4 &  \textbf{71.3} $\pm$ 0.5      & \underline{53.7} $\pm$ 0.5          & \textcolor{gray}{39.9} $\pm$ 0.2     & {69.8}         &  {5.1} & 4.0&  {5}            \\
{ITTA \cite{2023-DG-ITTA} (CVPR, 2023)}          & 57.7 $\pm$ 0.6        & \textbf{98.5} $\pm$ 0.1        & \textcolor{gray}{76.9} $\pm$ 0.6 & \textcolor{gray}{83.8} $\pm$ 0.3 &  \textcolor{gray}{62.0} $\pm$ 0.2      &  \textcolor{gray}{43.2} $\pm$ 0.5          & \textcolor{gray}{34.9} $\pm$ 0.1     & \textcolor{gray}{65.3}         &  \textcolor{gray}{14.1} & \textcolor{gray}{10.9}&  \textcolor{gray}{16}            \\
{SAGM \cite{2023-DG-SAGM} (CVPR, 2023)}          & 63.4 $\pm$ 1.2        & 98.0 $\pm$ 0.1        & \underline{79.9} $\pm$ 0.2 & \textcolor{gray}{85.8} $\pm$ 0.8 &  \textcolor{gray}{65.3} $\pm$ 0.5      &  \textcolor{gray}{50.8} $\pm$ 0.6          & \textcolor{gray}{38.5} $\pm$ 0.2     & {68.8}         &  {8.7} & 6.8&  \textcolor{gray}{11}            \\
{RDM \cite{2024-DG-RDM} (WACV, 2024)}          & \textcolor{gray}{57.5} $\pm$ 1.1        & 97.8 $\pm$ 0.0        & {78.4} $\pm$ 0.4 & \underline{87.2} $\pm$ 0.7 &  {67.3} $\pm$ 0.4 &  \textcolor{gray}{47.5} $\pm$ 1.0          & {41.4} $\pm$ 0.3     & \textcolor{gray}{68.2}         &  {8.7} & 6.9&  {6}            \\
\midrule
POGM (Ours)                      & 66.3 $\pm$ 1.2            & 97.9 $\pm$ 0.0            & \textbf{82.0} $\pm$ 0.1            & \textbf{88.4} $\pm$ 0.5            & \underline{70.0} $\pm$ 0.3            & \textbf{54.8} $\pm$ 0.5            & \underline{42.6} $\pm$ 0.2            & \textbf{71.7} &\textbf{2.1}	&\textbf{1.8} &	\textbf{2}
 \\ 
 \bottomrule
\end{tabular}}
\end{table*}

We conducted extensive experiments to validate our proposed POGM on DomainBed \cite{2021-DG-DomainBed}. 
Our evaluation encompasses not only synthetic datasets like Colored MNIST (CMNIST) \cite{2015-DG-ColoredMNIST} and Rotated MNIST (RMNIST) \cite{2015-DG-RotatedMNIST}, but also real-world multi-domain image classification datasets such as VLCS \cite{2011-DG-VLCS}, PACS \cite{2017-DG-PACS}, OfficeHome \cite{2017-DG-OfficeHome}, Terra Incognita \cite{2018-DG-TerraInCognita}, and DomainNet \cite{2019-DG-DomainNet}. To ensure a fair comparison, we enforce strict training conditions. All methods are trained using only 20 different hyperparameter configurations and the same number of steps. We then averaged the results over three trials. For a comprehensive understanding of our experimental setup, please refer to the detailed settings provided in Appendix~\ref{app:setings}. Tab.~\ref{tab:domainbed-benchmark} summarizes results on DomainBed using the \say{Test-domain} model selection: The validation set follows the same distribution as the test domain.

In DomainBed, ERM was carefully fine-tuned and therefore serves as a robust baseline. However, previous methods consistently fall short of achieving the highest scores across datasets. Specifically, invariant predictors like IRM and VREx, along with gradient-masking approaches such as AND-mask, demonstrate poor performance on real datasets. Furthermore, CORAL not only underperforms compared to ERM on Terra Incognita but, more crucially, it fails to identify correlation shifts on CMNIST. This is attributed to feature-based methods neglecting label considerations.

As observed from Tab.~\ref{tab:domainbed-benchmark}, our proposed method efficiently tackles correlation and diversity shifts. Besides the Fishr, which shows the competitive results of our work, our POGM systematically performs better than ERM on all real datasets: the differences are over standard errors on VLCS ($82.0\%$ vs. $77.6\%$), PACS ($88.4\%$ vs. $86.7\%$), OfficeHome ($70.0\%$ vs. $66.4\%$), and on the large-scale Terra Inc. ($54.8\%$ vs. $53.0\%$), and DomainNet ($42.6\%$ vs. $41.3\%$). However, on synthetic data, our proposed POGM does not exhibit significant improvement over other baseline methods. Notably, POGM performs notably worse on CMNIST, which will be discussed further in the subsequent section. In summary, despite the decrease in synthetic datasets, POGM shows competitive results vs. Fish, and Fishr outperforms the ERM on all challenging datasets. Significantly, POGM consistently achieves top results on real datasets, i.e., ranking in the top 1 and 2 positions on benchmark tests.

\subsection{Numerical Studies}
\subsubsection{Invariant Gradient Properties}\label{sec:invariant-properties}
\begin{figure}[!ht]
    \begin{subfigure}[b]{0.24\textwidth}
        \includegraphics[width=\textwidth]{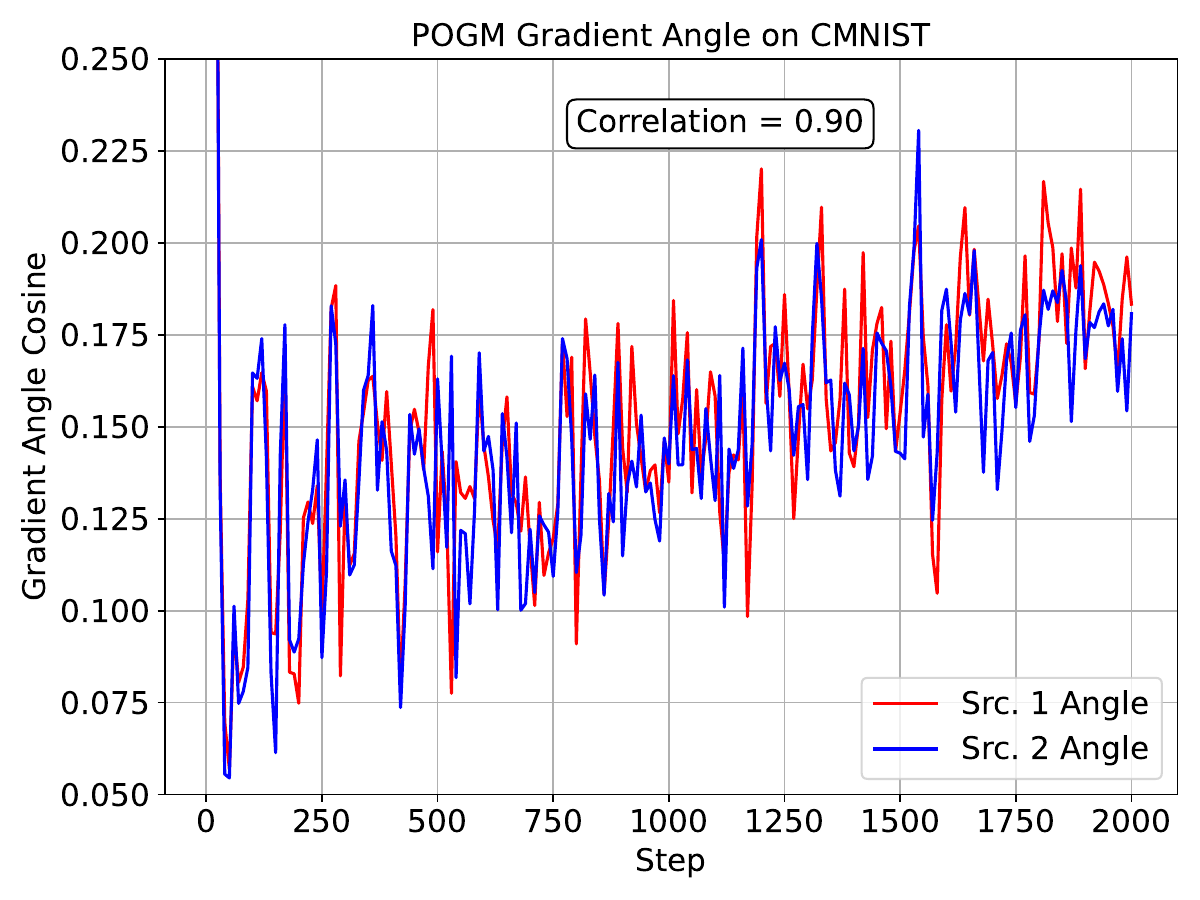}
        \caption{}
        \vspace{0.3cm}
        \label{fig:cmnist-pogm-angle}
    \end{subfigure}%
    \begin{subfigure}[b]{0.24\textwidth}
        \includegraphics[width=\textwidth]{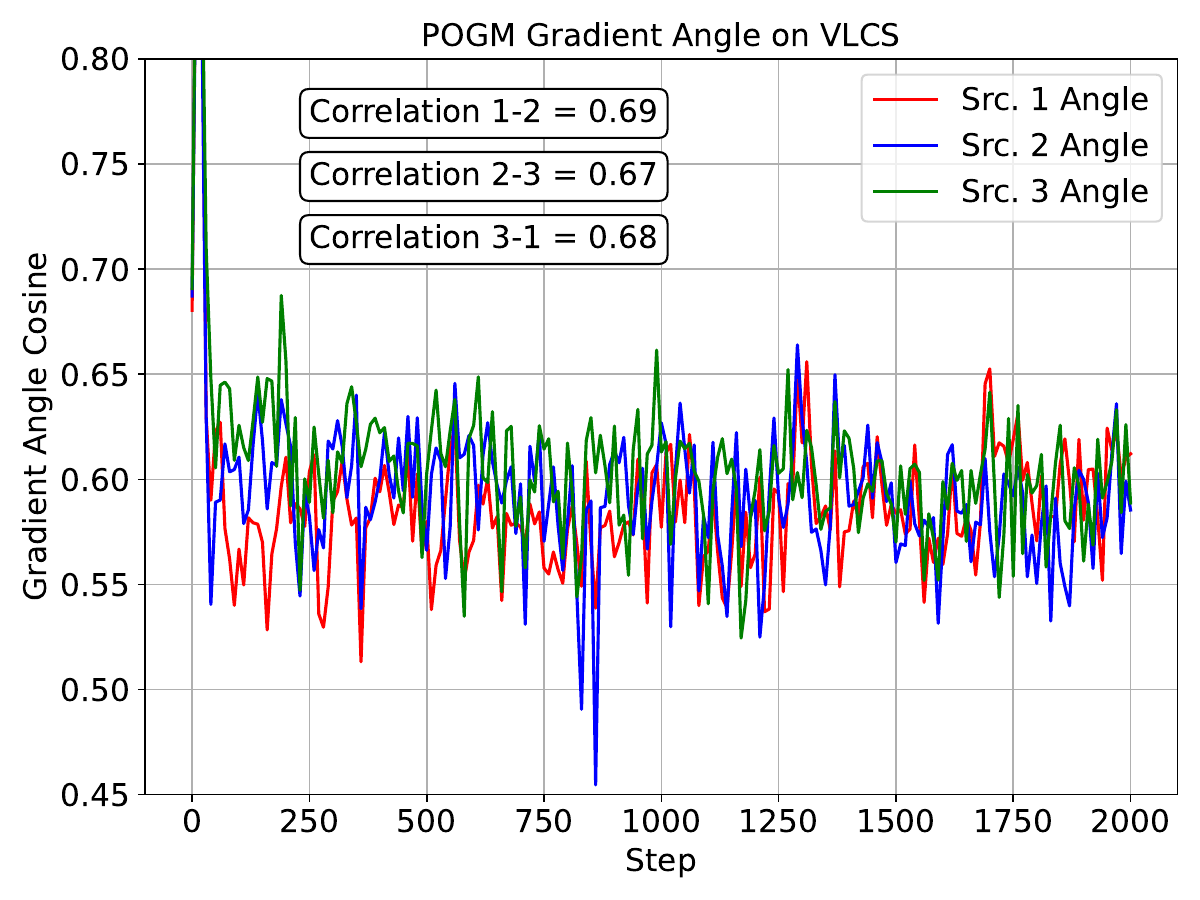}
        \caption{}  
        \vspace{0.3cm}
        \label{fig:vlcs-pogm-angle}
    \end{subfigure}
    \begin{subfigure}[b]{0.24\textwidth}
        \includegraphics[width=\textwidth]{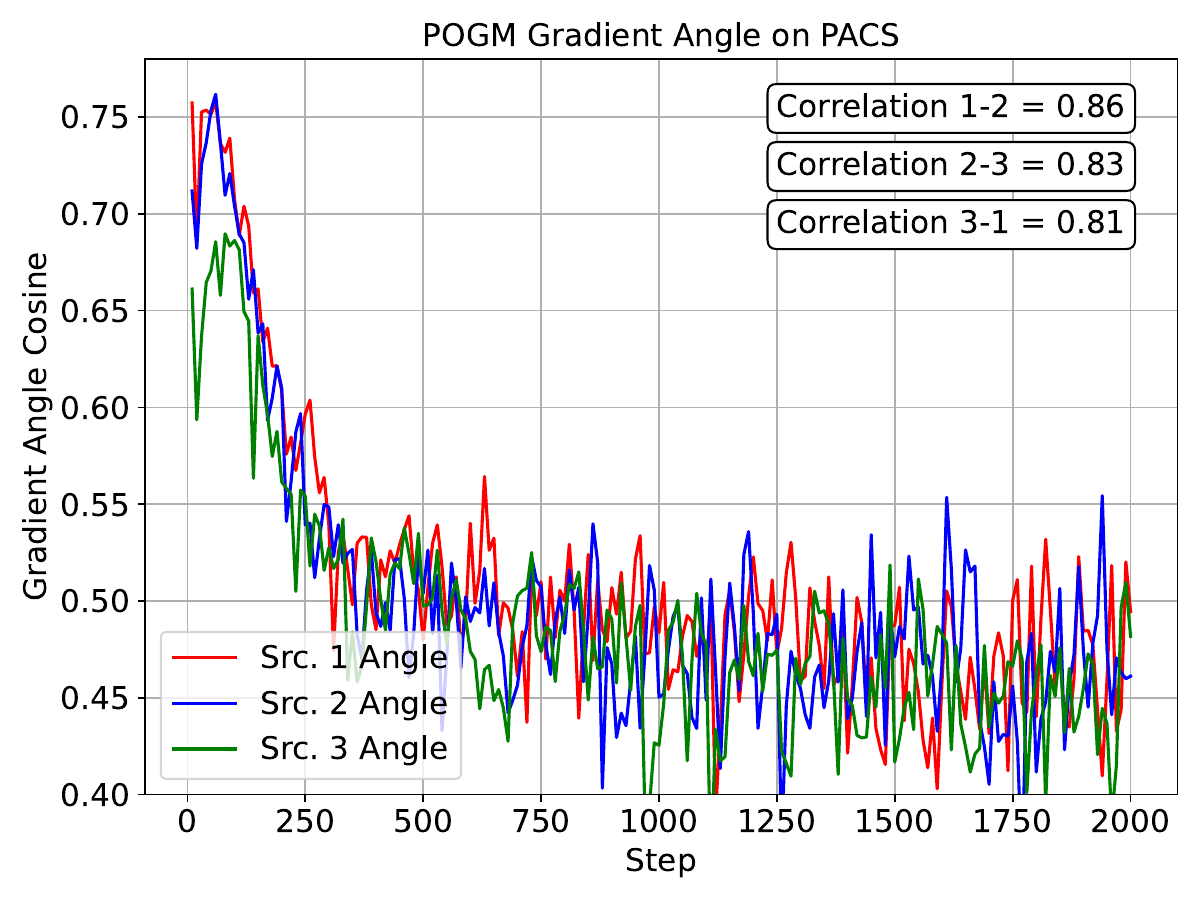}
        \caption{}
        \label{fig:pacs-pogm-angle}
    \end{subfigure}%
    \begin{subfigure}[b]{0.24\textwidth}
        \includegraphics[width=\textwidth]{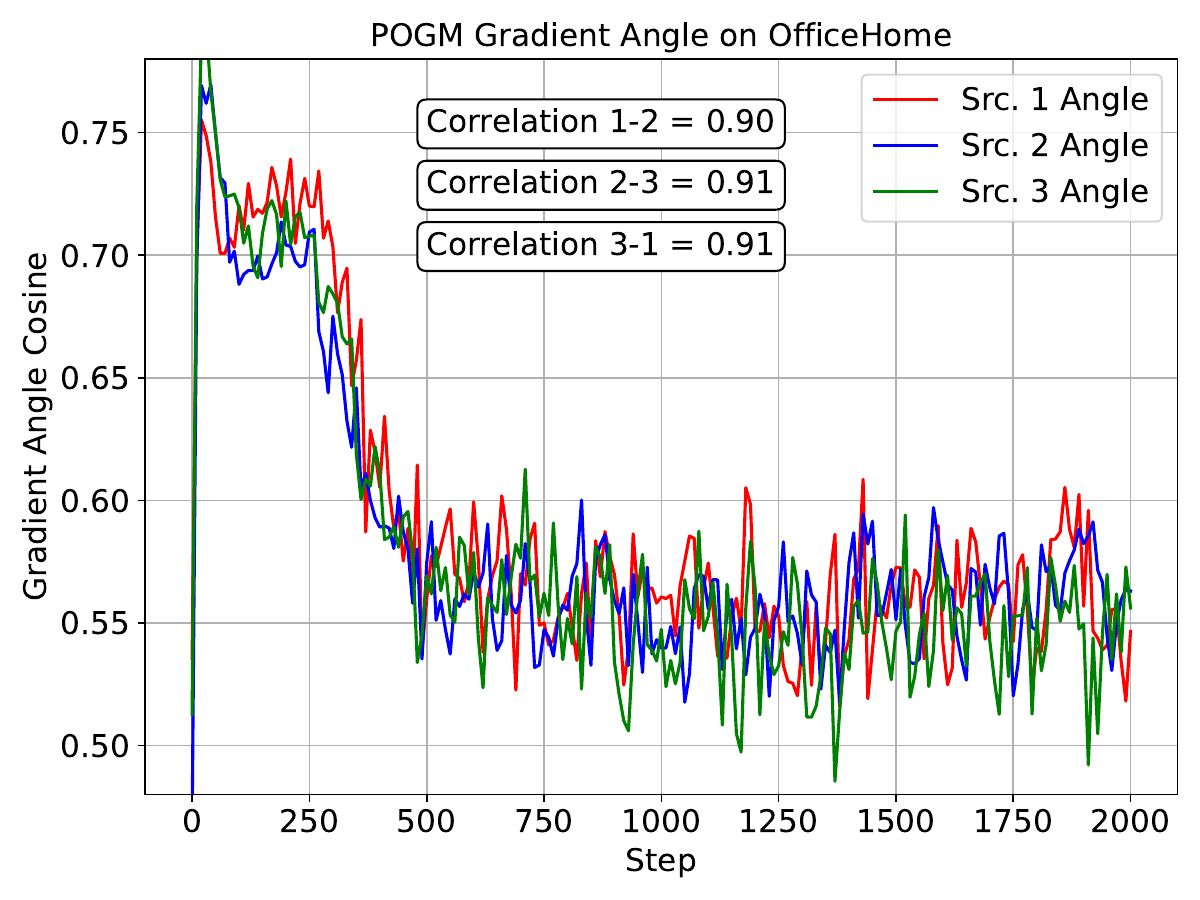}
        \caption{}  
        \label{fig:officehome-pogm-angle}
    \end{subfigure}
    \vspace{0.3cm}
    \caption{The correlation of gradient direction of POGM on CMNIST, VLCS, PACS, and OfficeHome. Compared to Fish and Fishr, POGM represents a higher correlation of gradient pairs among domains.}
    \vspace{0.3cm}
    \label{fig:pogm-invariant-angle}
\end{figure}

Fig.~\ref{fig:pogm-invariant-angle} illustrates the invariant gradient properties of POGM. POGM shows a stronger correlation between the two domain-specific gradient angles compared to ERM, Fish (refer to Fig.~\ref{fig:fish-erm-angle}).
Furthermore, every pair-wise gradient angle has a smaller gap from each other; thus, the gradient of POGM shows better invariant properties than that of Fish and Fishr.
The correlation implies that the angles of two gradient directions change at the same rate as the POGM, indicating invariant properties. 

\subsubsection{Domain Specific Gradient Angle and Norm Difference}\label{sec:norm-angle}
\begin{figure}[!ht]
    \begin{subfigure}[b]{0.24\textwidth}
        \includegraphics[width=\textwidth]{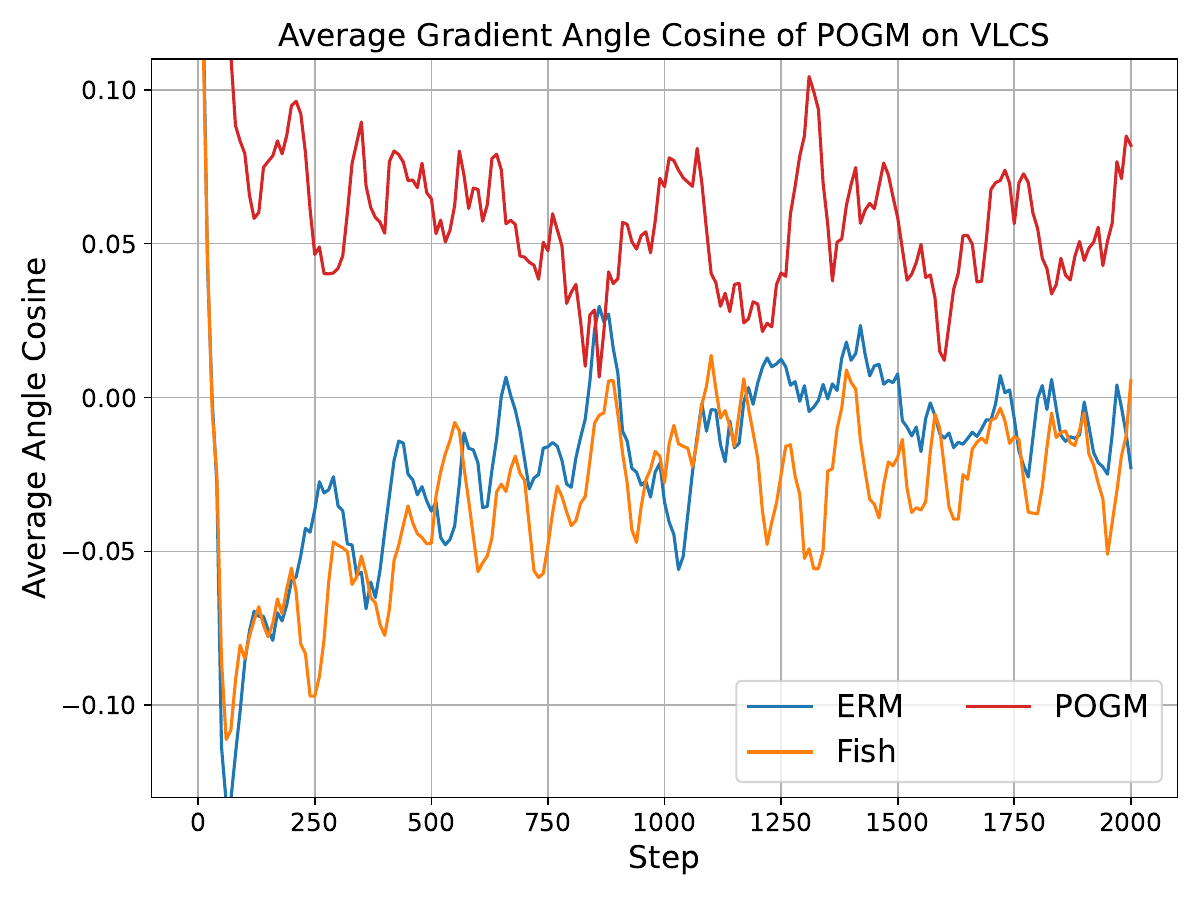}
        \caption{}
        \vspace{0.3cm}
        \label{fig:vlcs-avg-angle}
    \end{subfigure}%
    \begin{subfigure}[b]{0.24\textwidth}
        \includegraphics[width=\textwidth]{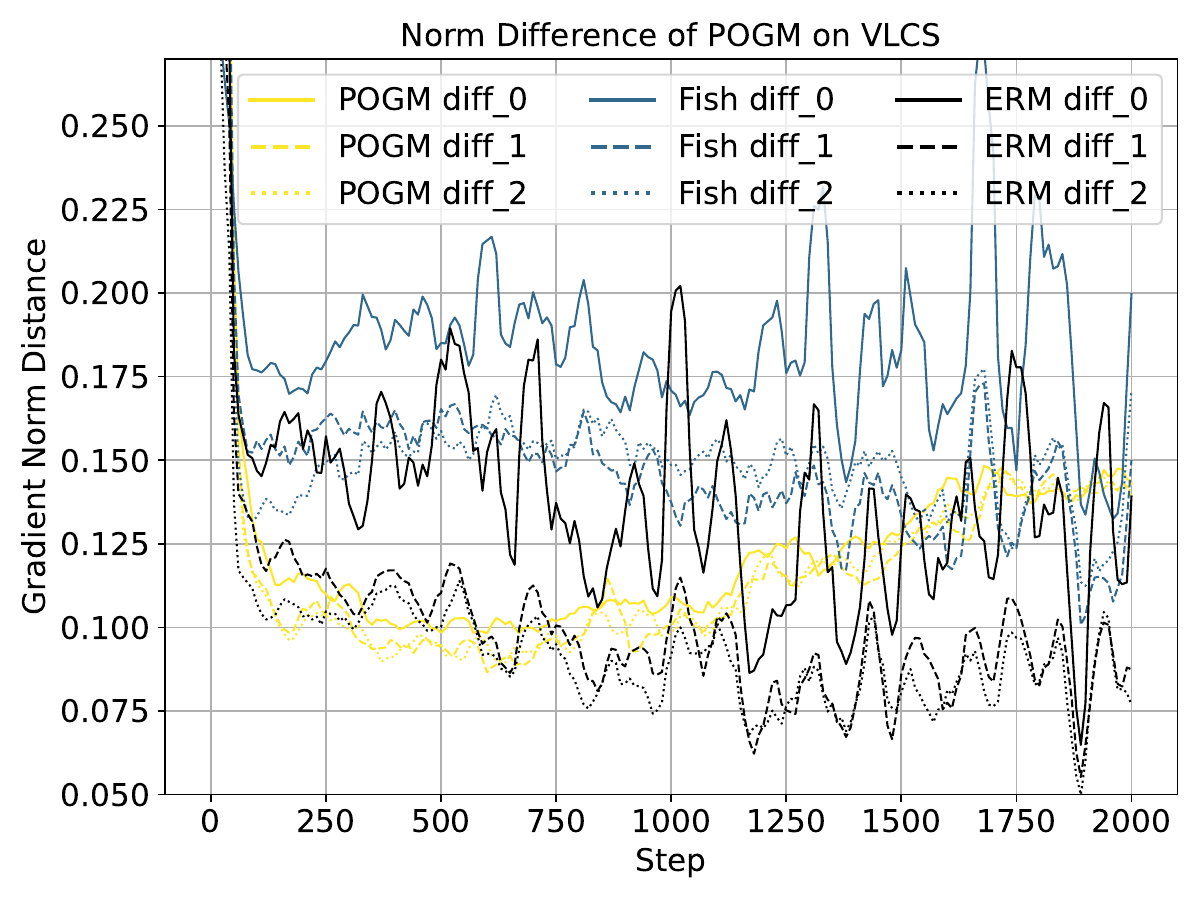}
        \caption{}  
        \vspace{0.3cm}
        \label{fig:vlcs-diff-norm}
    \end{subfigure}
    \begin{subfigure}[b]{0.24\textwidth}
        \includegraphics[width=\textwidth]{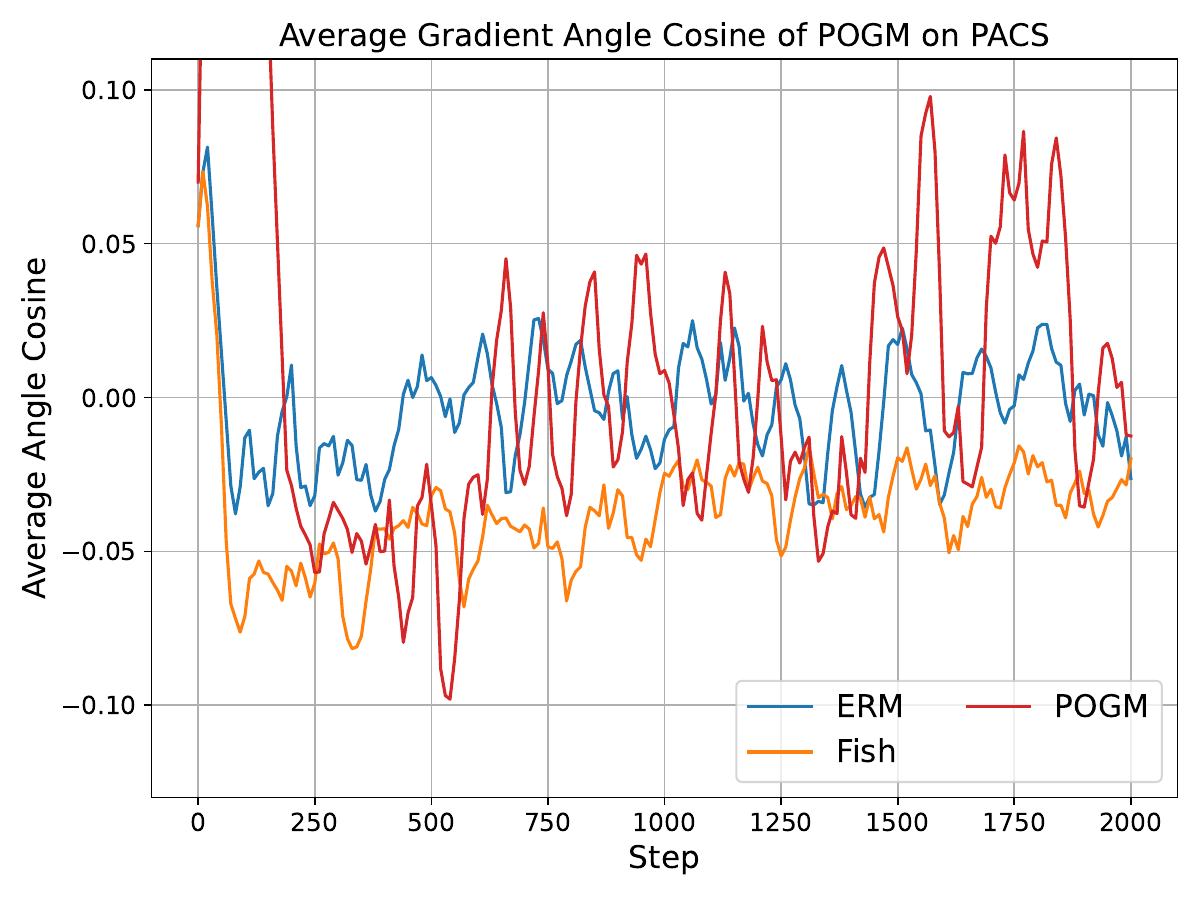}
        \caption{}
        \label{fig:pacs-avg-angle}
    \end{subfigure}%
    \begin{subfigure}[b]{0.24\textwidth}
        \includegraphics[width=\textwidth]{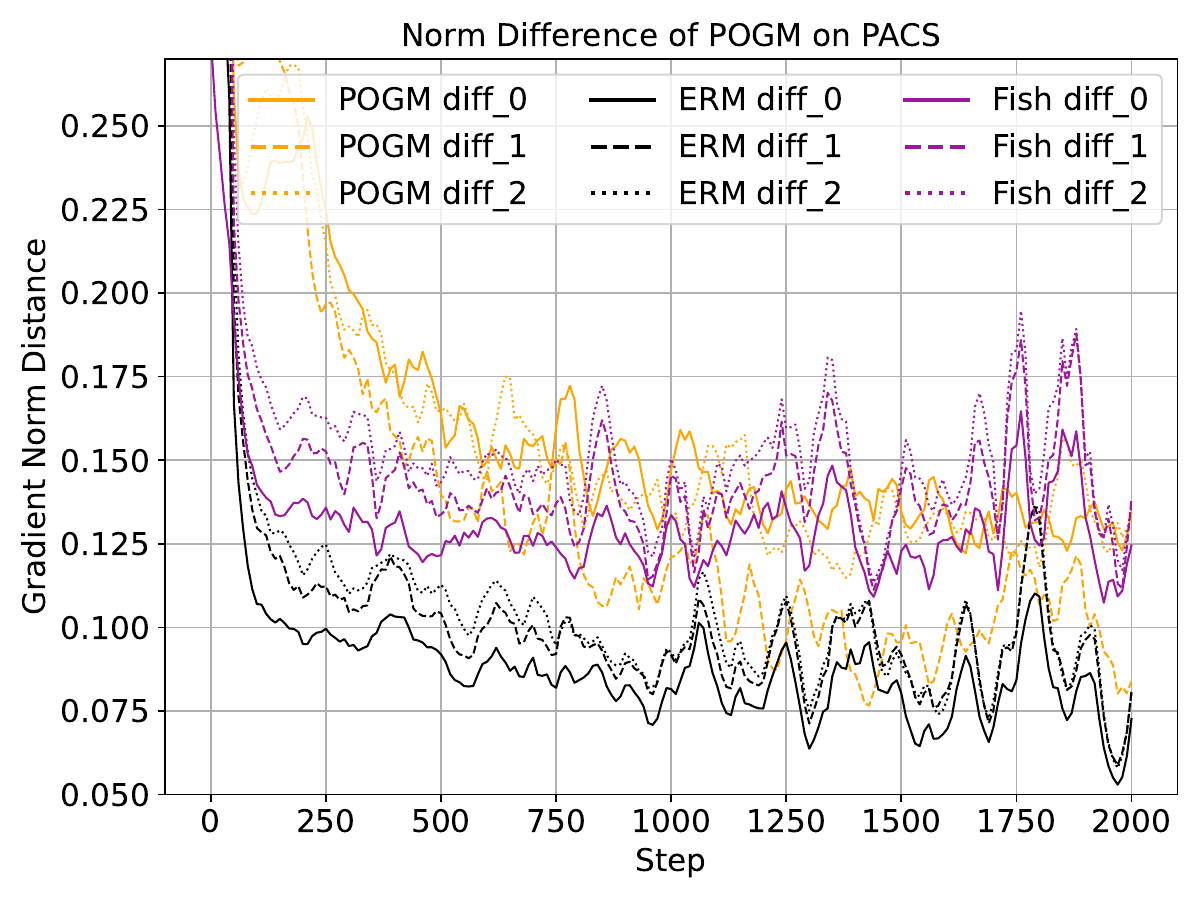}
        \caption{}  
        \label{fig:pacs-diff-norm}
    \end{subfigure}
    \vspace{0.3cm}
    \caption{The gradients, angles, and the norm difference of ERM, Fish, Fishr, POGM on VLCS, PACS.}
    \vspace{0.5cm}
    \label{fig:grad-norm}
\end{figure}

Besides the gradient invariant properties, we consider the average angle cosine with the domain-specific gradients $\cos(\theta^{(r)}_\textrm{GIP-C},\theta^{(r)}_i)$ and the norm distance with the domain-specific models $\Vert \theta^{(r)}_\textrm{GIP-C} - \theta^{(r)}_i\Vert$. Suppose the gradients' angles and the norm distances are small. In that case, the generalization gap between the trained model and the domain-specific model is better, thus improving the validation set on the source domains. The results in Fig.~\ref{fig:grad-norm} demonstrate that POGM outperforms Fish in retaining the performance on source domains while still improving the results on target domains. Additionally, models tend to perform better when the average gradient angle cosine exceeds $0$, indicating most gradient conflicts \cite{2023-MTL-recon} become insignificant (i.e., the angles become less than 90 degrees).

\subsection{Stable training behavior of POGM}\label{sec:consistent-pogm}

\begin{figure}[!ht]
    \begin{subfigure}[b]{0.25\textwidth}
        \includegraphics[width=\textwidth]{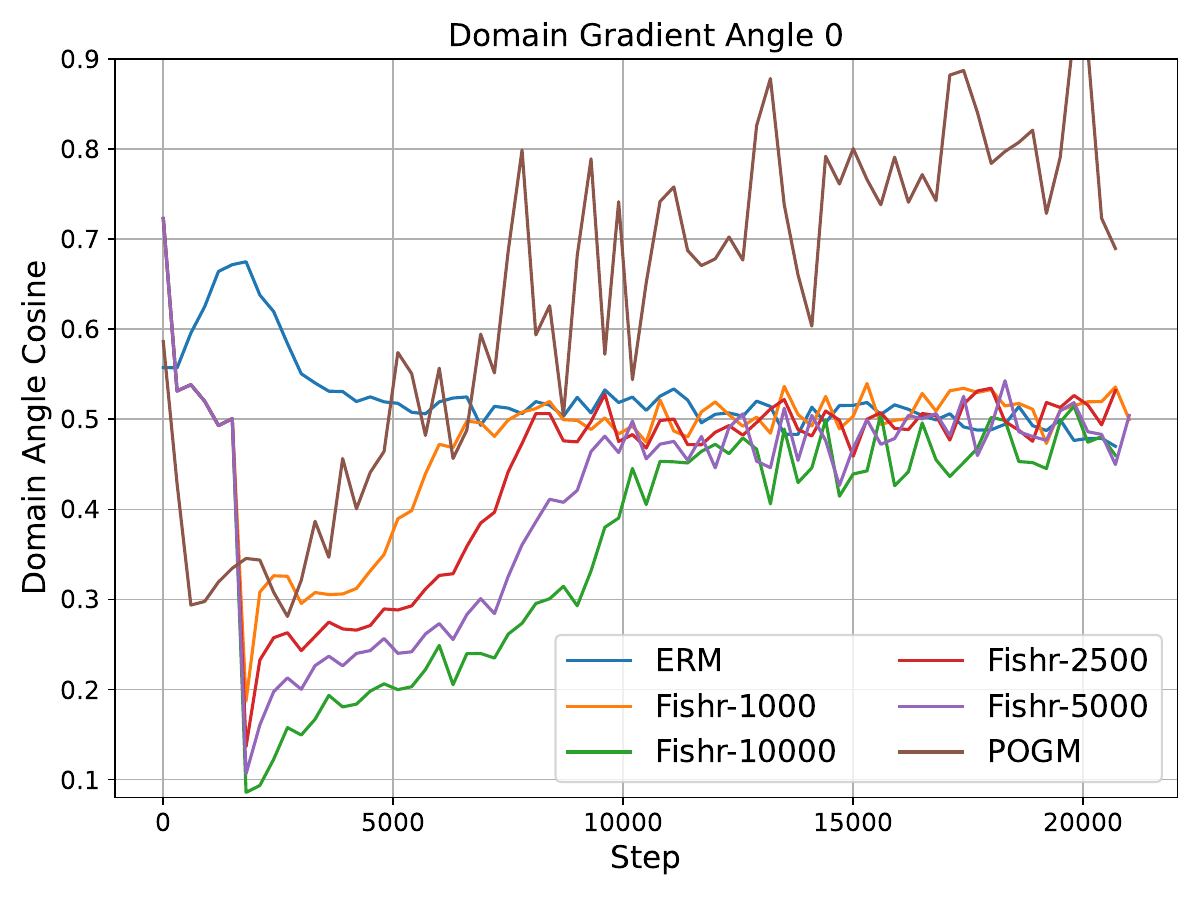}
        \caption{}
        \vspace{0.3cm}
        \label{fig:POGM-angle0}
    \end{subfigure}%
    \begin{subfigure}[b]{0.25\textwidth}
        \includegraphics[width=\textwidth]{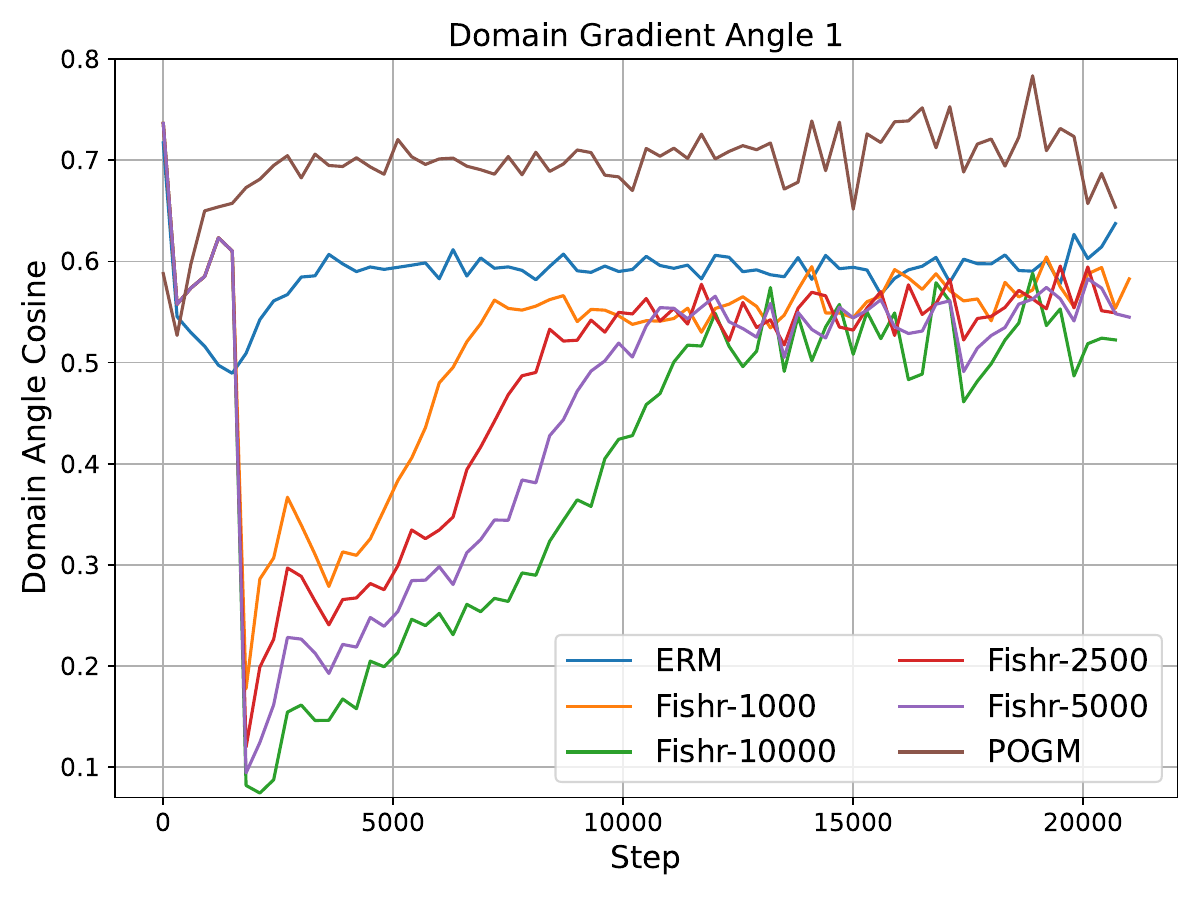}
        \caption{}        
        \vspace{0.3cm}
        \label{fig:POGM-angle1}
    \end{subfigure}
    \begin{subfigure}[b]{0.25\textwidth}
        \includegraphics[width=\textwidth]{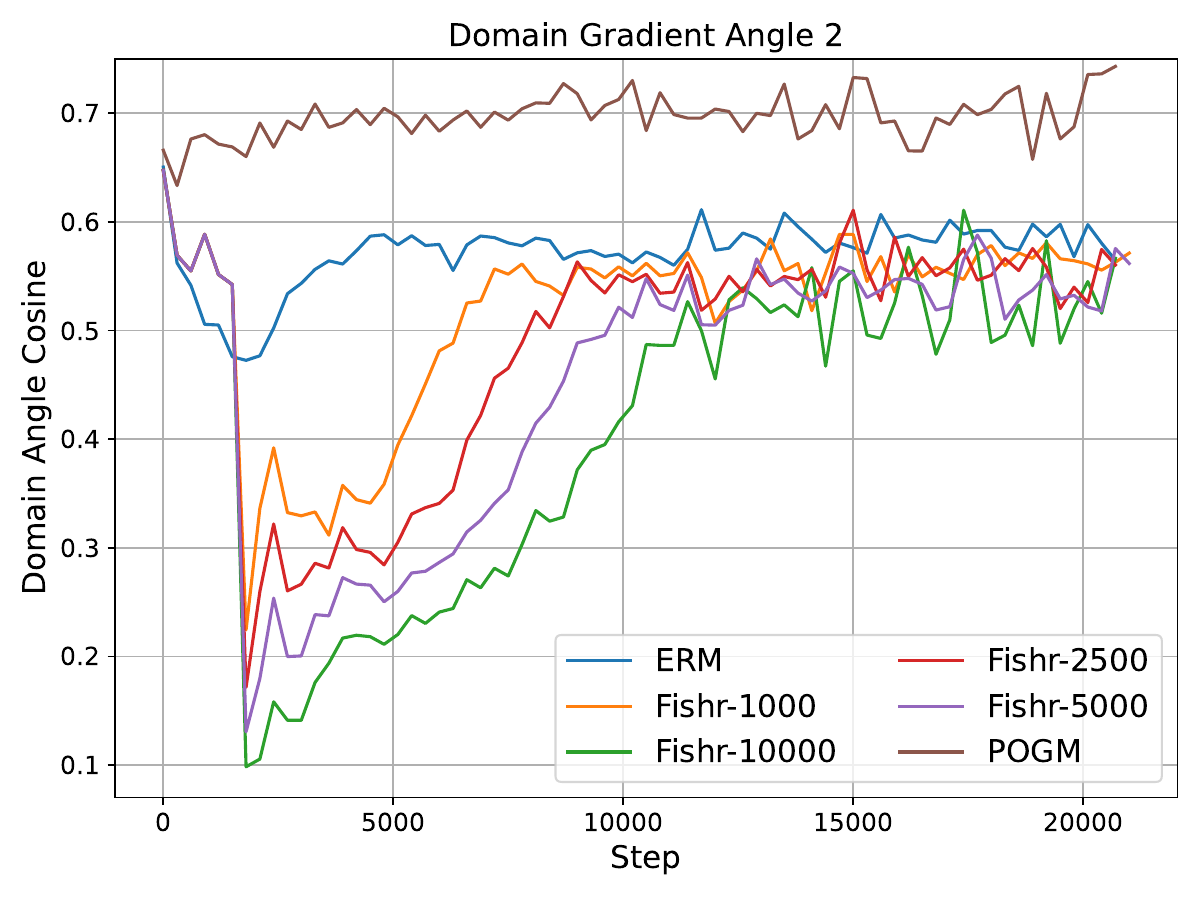}
        \caption{}
        \label{fig:POGM-angle2}
    \end{subfigure}%
    \begin{subfigure}[b]{0.25\textwidth}
        \includegraphics[width=\textwidth]{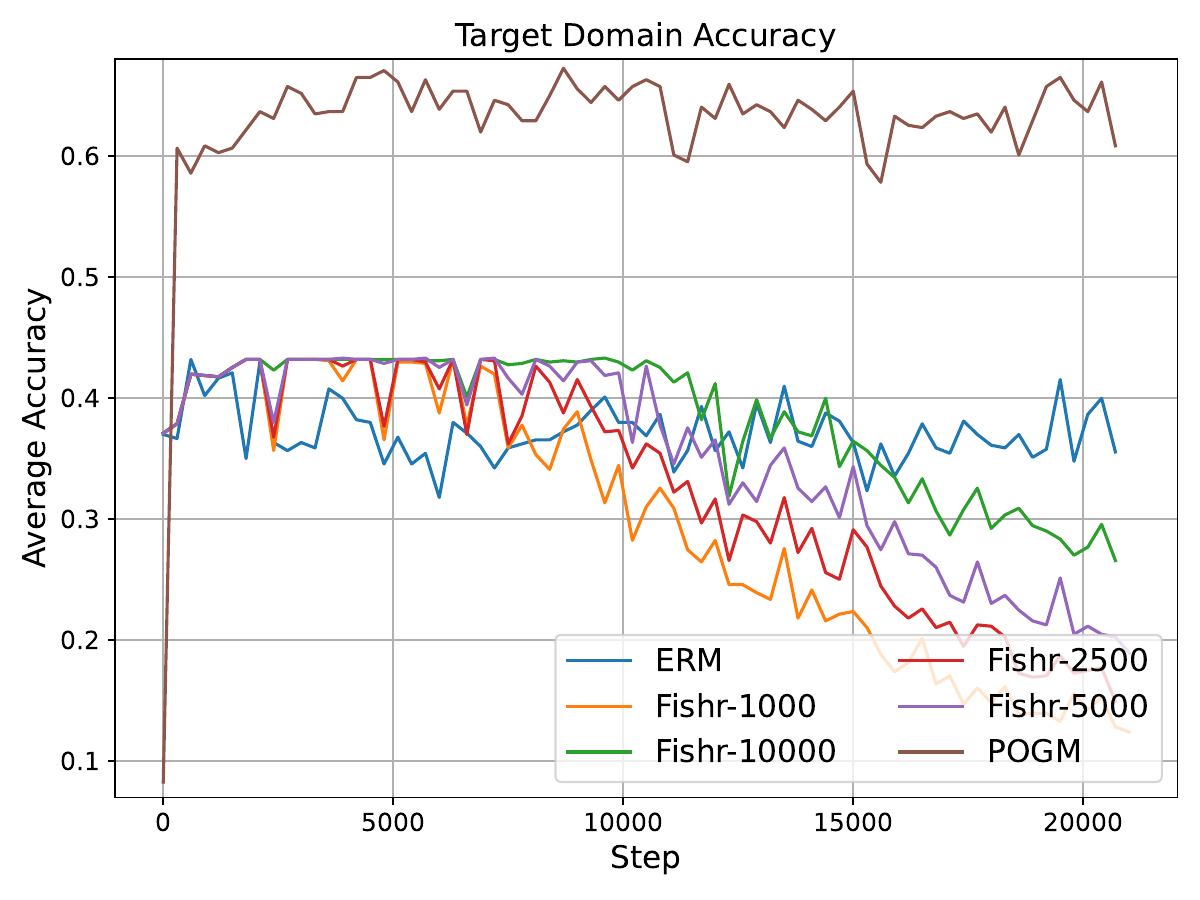}
        \caption{}
        \label{fig:POGM-TestAcc}
    \end{subfigure}%
    \vspace{0.3cm}
    \caption{Invariant properties of POGM vs. ERM and Fishr.}
    \label{fig:invariant-properties}
    \vspace{0.5cm}
\end{figure}
Despite the robust results reported in the benchmarks, Fishr has a drawback that impedes the convergence rate of Fishr in the initial rounds and can be further improved. Our introduced POGM can avoid the drawback of Fishr. Fig.\ref{fig:invariant-properties} illustrates the performance of POGM over Fishr and ERM in the VLCS dataset. In Fishr, by minimizing the MSE among gradient variance and mean, the gradient angles of Fishr do not achieve the optimal results at the initial phases (from step $500$ to $5000$ in figures~\ref{fig:POGM-angle0}, \ref{fig:POGM-angle1}, and \ref{fig:POGM-angle2}). In the latter stages, the Fishr angles become similar to those of ERM. 

In contrast, by following GIP maximization proposed by \cite{2022-DG-Fish}, the angles between the learned gradient and other domains are optimized directly, resulting in a significant improvement in POGM over ERM and Fishr (see Fig.~\ref{fig:POGM-TestAcc}).

\subsection{Integratability}\label{sec:integration}
From Tab.~\ref{tab:integratation}, POGM demonstrates strong performance when combined with representation augmentation and mixing methods (e.g., Mixup and CIRL) in significant domain shifts (i.e., CMNIST). This is because representation mixing reduces domain shifts among different domains \cite{2022-DG-CMixup}. 

\begin{table}[ht]
\centering
\centering
\caption{The integrability of POGM}
\small
\begin{tabular}{lccc}
\toprule
\textbf{Dataset} & CMNIST & PACS & VLCS \\
\midrule
POGM       & 66.3 $\pm$ 1.2 & {88.4} $\pm$ 0.5 & {70.0} $\pm$ 0.3 \\
POGM + Mixup  & 69.5 $\pm$ 0.6 & 89.1 $\pm$ 0.6 & {70.8} $\pm$ 0.5 \\
POGM + Data Aug. & 67.1 $\pm$ 1.1 & {88.6} $\pm$ 0.5 & {71.2} $\pm$ 0.4 \\
POGM + CIRL   & \textbf{71.4} $\pm$ 0.7 & {90.1} $\pm$ 0.4 &  {70.6} $\pm$ 0.5 \\
POGM + SWAD  & 68.1 $\pm$ 0.5 & \textbf{91.2} $\pm$ 0.3 & \textbf{72.6} $\pm$ 0.2 \\
\bottomrule
\end{tabular}
\label{tab:integratation}
\end{table}

\begin{table*}[!h]
\caption{Ablation test on varying meta update learning rate}
\label{tab:meta-lr}
\centering
\adjustbox{max width=\textwidth}{%
\begin{tabular}{clcccccccc}
\hline
\multicolumn{1}{l}{\textbf{Model selection}} & \textbf{$\alpha$} & \textbf{CMNIST} & \textbf{RMNIST} & \textbf{VLCS} & \textbf{PACS} & \textbf{OfficeHome} & \textbf{TerraInc} & \textbf{DomainNet} & \textbf{Avg} \\ \hline
\multirow{3}{*}{Testing-domain} 
 & 0.01 & 65.2 $\pm$ 0.9            & 94.7 $\pm$ 0.3            & 81.3 $\pm$ 0.3            & 87.6 $\pm$ 0.9            & 68.7 $\pm$ 0.5            & 50.4 $\pm$ 0.9            & 30.3 $\pm$ 2.0            & 68.3 \\
 & 0.1 & 62.0 $\pm$ 0.9            & 93.4 $\pm$ 0.9            & 81.3 $\pm$ 0.2            & 88.5 $\pm$ 0.5            & 69.4 $\pm$ 0.6            & 51.4 $\pm$ 0.5            & 34.2 $\pm$ 1.1            & 68.6 \\
 & 0.5 & 64.7 $\pm$ 1.2            & 97.8 $\pm$ 0.1            & 80.5 $\pm$ 0.4            & 87.8 $\pm$ 0.3            & 68.6 $\pm$ 0.4            & 53.9 $\pm$ 0.1            & 40.0 $\pm$ 0.4            & 70.5   \\ \hline
\multirow{3}{*}{Training-domain} 
 & 0.01 & 51.7 $\pm$ 0.2            & 94.0 $\pm$ 0.2            & 79.4 $\pm$ 0.3            & 82.8 $\pm$ 0.7            & 67.8 $\pm$ 0.5            & 46.0 $\pm$ 1.1            & 30.3 $\pm$ 2.0            & 64.6 \\
 & 0.1 & 51.1 $\pm$ 0.2            & 93.2 $\pm$ 0.8            & 79.7 $\pm$ 0.6            & 86.4 $\pm$ 0.9            & 69.2 $\pm$ 0.7            & 46.0 $\pm$ 1.7            & 34.2 $\pm$ 1.1            & 65.7 \\
 & 0.5 & 51.4 $\pm$ 0.2            & 97.7 $\pm$ 0.1            & 79.4 $\pm$ 0.3            & 85.8 $\pm$ 0.6            & 67.8 $\pm$ 1.0            & 46.5 $\pm$ 0.4            & 40.0 $\pm$ 0.4            & 66.9   \\ \hline
\end{tabular}}
\end{table*}

\begin{table*}[!h]
\caption{Ablation test on varying domain-specific training iterations}
\label{tab:kappa}
\centering
\adjustbox{max width=\textwidth}{%
\begin{tabular}{clcccccccc}
\hline
\multicolumn{1}{l}{\textbf{Model selection}} & \textbf{$E$} & \textbf{CMNIST} & \textbf{RMNIST} & \textbf{VLCS} & \textbf{PACS} & \textbf{OfficeHome} & \textbf{TerraInc} & \textbf{DomainNet} & \textbf{Avg} \\ \hline
\multirow{3}{*}{Testing-domain} 
 & 1 & 63.8 $\pm$ 1.2            & 97.3 $\pm$ 0.3            & 79.7 $\pm$ 0.6            & 87.3 $\pm$ 0.4            & 67.5 $\pm$ 0.4            & 54.0 $\pm$ 0.6            & 40.0 $\pm$ 0.4            & 69.9 \\
 & 5 & 60.9 $\pm$ 1.3            & 95.8 $\pm$ 0.7            & 81.0 $\pm$ 0.1            & 87.5 $\pm$ 0.9            & 69.0 $\pm$ 1.0            & 49.9 $\pm$ 1.0            & 33.1 $\pm$ 2.0            & 68.2  \\
 & 10 & 60.3 $\pm$ 1.2            & 97.5 $\pm$ 0.1            & 81.5 $\pm$ 0.4            & 88.2 $\pm$ 0.4            & 69.7 $\pm$ 0.2            & 52.3 $\pm$ 0.9            & 32.1 $\pm$ 0.6            & 68.8  \\ \hline
\multirow{3}{*}{Training-domain} & 1 & 51.6 $\pm$ 0.2            & 97.2 $\pm$ 0.3            & 78.8 $\pm$ 0.6            & 84.0 $\pm$ 0.3            & 67.0 $\pm$ 0.6            & 46.0 $\pm$ 0.4            & 40.0 $\pm$ 0.4            & 66.4  \\
 & 5 & 51.3 $\pm$ 0.1            & 95.6 $\pm$ 0.7            & 80.2 $\pm$ 0.3            & 86.5 $\pm$ 0.8            & 68.7 $\pm$ 1.0            & 44.9 $\pm$ 1.2            & 33.1 $\pm$ 2.0            & 65.7   \\
 & 10 & 50.4 $\pm$ 0.2            & 97.3 $\pm$ 0.2            & 79.1 $\pm$ 0.2            & 86.0 $\pm$ 0.3            & 69.0 $\pm$ 0.2            & 47.1 $\pm$ 0.9            & 31.9 $\pm$ 0.7            & 65.8   \\ \hline
\end{tabular}}
\end{table*}

\begin{table*}[!h]
\caption{Ablation test in varying searching radius}
\label{tab:radius}
\centering
\adjustbox{max width=\textwidth}{%
\begin{tabular}{clcccccccc}
\hline
\multicolumn{1}{l}{\textbf{Model selection}} & \textbf{$\kappa$} & \textbf{CMNIST} & \textbf{RMNIST} & \textbf{VLCS} & \textbf{PACS} & \textbf{OfficeHome} & \textbf{TerraInc} & \textbf{DomainNet} & \textbf{Avg} \\ \hline
\multirow{3}{*}{Testing-domain} 
 & 0.05 & 62.0 $\pm$ 1.4            & 96.3 $\pm$ 0.5            & 80.4 $\pm$ 0.8            & 85.7 $\pm$ 1.5            & 68.3 $\pm$ 1.1            & 49.9 $\pm$ 0.9            & 28.1 $\pm$ 3.6            & 67.2  \\
 & 0.1 & 64.5 $\pm$ 1.6            & 97.4 $\pm$ 0.2            & 81.2 $\pm$ 0.4            & 88.2 $\pm$ 0.8            & 68.7 $\pm$ 0.6            & 52.9 $\pm$ 0.6            & 31.6 $\pm$ 0.9            & 69.2   \\
 & 0.5 & 63.8 $\pm$ 1.0            & 97.2 $\pm$ 0.3            & 82.0 $\pm$ 0.1            & 88.3 $\pm$ 0.4            & 69.5 $\pm$ 0.2            & 53.9 $\pm$ 0.1            & 40.2 $\pm$ 0.3            & 70.7 \\ \hline
\multirow{3}{*}{Training-domain} 
 & 0.05 & 51.4 $\pm$ 0.1            & 96.1 $\pm$ 0.7            & 79.0 $\pm$ 0.8            & 84.4 $\pm$ 1.6            & 68.1 $\pm$ 1.3            & 45.3 $\pm$ 1.0            & 28.1 $\pm$ 3.6            & 64.6  \\
 & 0.1 & 51.3 $\pm$ 0.1            & 97.2 $\pm$ 0.3            & 79.5 $\pm$ 0.3            & 84.7 $\pm$ 0.8            & 67.4 $\pm$ 0.6            & 45.4 $\pm$ 1.5            & 31.5 $\pm$ 0.8            & 65.3 \\
 & 0.5 & 51.0 $\pm$ 0.5            & 97.2 $\pm$ 0.3            & 78.7 $\pm$ 0.3            & 85.8 $\pm$ 0.4            & 68.6 $\pm$ 0.4            & 45.8 $\pm$ 0.4            & 40.1 $\pm$ 0.3            & 66.7   \\ \hline
\end{tabular}}
\end{table*}

\subsection{Ablation Studies}
\subsection{Different Meta Update Learning Rate}

The ablation test of meta update learning rate is demonstrated as in Tab.~\ref{tab:meta-lr}. On the synthetic dataset RMNIST, due to the high correlation among domains, the DG problem tends to be simple. Thus, by choosing a high learning rate, we can easily achieve the optimal state. 

When dealing with the more challenging dataset (i.e., CMNIST with low correlation among domains and real datasets such as VLCS, PACS, and OfficeHome), the low learning rate appears to be the efficient setting. 

However, in real-world datasets with high dimensionality, and due to the large size of the learning model (i.e., ResNet-50), the loss landscape has significantly sharp minimizers. As a result, choosing a large meta-learning rate is efficient in these datasets.

\subsection{Different Domain-specific Training Iterations}

Tab.~\ref{tab:kappa} demonstrates the ablation test on different domain-specific training iterations on $7$ evaluating datasets (i.e., RMNIST, CMNIST, VLCS, PACS, OfficeHome, Terra Incognita, and DomainNet). The results differ depending on the dataset characteristics. For synthetic datasets like RMNIST and CMNIST, where the data is simpler, gradient trajectories are easily estimated. Thus, using a low number of domain-specific training iterations between each meta-update round (e.g., 1 round) yields optimal settings. However, for real-world datasets such as VLCS, PACS, OfficeHome, Terra Incognita, and DomainNet, employing a larger number of rounds leads to improved training outcomes.

\subsection{Different Searching Hypersphere Radius}

Tab.~\ref{tab:radius} demonstrates the ablation test on different searching hypersphere radius $\kappa$ on 7 evaluating datasets (i.e., RMNIST, CMNIST, VLCS, PACS, OfficeHome, Terra Incognita, DomainNet). Across almost all datasets, POGM performs well when the available search radius is set to $\kappa\geq 0.5$. We can explain this problem as follows. Because of significant gradient divergence across domains, the search space expands accordingly. Therefore, the optimal solution is significantly different from the ERM solutions, likely due to gradient divergence, which expands the search space. Besides, in the synthetic dataset RMNIST, the difference among domains is not significant, which makes the search space become small. Therefore, the optimal solution is $\kappa=0.1$.


\section{Limitations and Future Works}
While the proposed method benefits from computational efficiency, achieved through the use of surrogate learnable parameters that eliminate the need for second-order derivatives during gradient matching, it still presents certain limitations. Specifically, the current implementation may encounter out-of-memory issues when applied to very large models. This limitation arises from the heavy reliance on matrix multiplication operations within the algorithm. Moreover, the learnable parameter associated with each gradient in the domain-wise gradient matching is currently limited to a scalar value. As a result, it can only scale the magnitude of the gradient without influencing its direction. We hypothesize that the performance of POGM could be further enhanced by designing a set of learnable parameters capable of rotating gradient directions under appropriate constraints.

Given that POGM relies solely on gradients without requiring direct access to data, it shows promising potential for applications in distributed learning and multi-agent systems, where data sharing among distributed devices is often restricted.

\section{Conclusion}
In our paper, we tackled the challenge of out-of-distribution generalization. We conducted thorough empirical analyses on two state-of-the-art gradient-based methods, Fish \cite{2022-DG-Fish} and Fishr \cite{2022-DG-Fishr}, revealing issue of gradient fluctuation. These issues hinder both Fishr and Fish from consistently achieving peak performance. Building on these observations, we propose a novel approach that incorporates GIP from Fish and introduces a generalized regularization, called GIP-C, to ensure stability. We employ meta-learning to separate the domain-specific optimization stage from the GIP optimization phase, allowing for a Hessian-free approximation of our GIP-C optimization problem. Our experiments, which are reproducible with our open-source implementation, demonstrate that POGM delivers competitive results compared to Fishr and outperforms other baseline methods across various popular datasets. We anticipate that our learning architecture will pave the way for gradient-based out-of-distribution generalization without the need for Hessian approximation.

\bibliography{main}






\newpage
\onecolumn
\appendix
\section{Experimental Settings}\label{app:setings}
\subsection{Datasets}
\paragraph{Rotated MNIST} \cite{2015-DG-ColoredMNIST} consists of $10000$ digits in MNIST with different rotated angle $d$ such that each domain is determined by the degree $d\in \{0,15,30,45,60,75\}$.

\paragraph{Colored MNIST} \cite{2021-DG-DomainBed} consists of $10000$ digits in MNIST with different rotated angle $d$ such that each domain is determined by the different color set.

\paragraph{PACS} \cite{2017-DG-PACS} includes $9991$ images with $7$ classes $y\in$ \{dog, elephant, giraffe, guitar, horse, house, person\} from $4$ domains $d\in$ \{art, cartoons, photos, sketches\}.

\paragraph{VLCS} \cite{2011-DG-VLCS} is composed of $10729$ images, $5$ classes $y\in$ \{bird, car, chair, dog, person\} from domains $d\in$ \{Caltech101, LabelMe, SUN09, VOC2017\}.

\paragraph{OfficeHome} \cite{2017-DG-OfficeHome} includes $15500$ images from $65$ categories from four domains $d\in$ \{Art, Clipart, Product, and Real-World\} with various categories of objects commonly found in office and home settings. It's widely used for domain adaptation and generalization tasks.

\paragraph{TerraIncognita} \cite{2018-DG-TerraInCognita} contains $24330$ images captured by satellites and aerial platforms, representing different terrains like forests, deserts, and urban areas.

\paragraph{DomainNet} \cite{2019-DG-DomainNet} is a large-scale dataset with $0.6$ million images, which are divided into 345 classes for DG research, featuring images from six domains $d\in$ \{Clipart, Infograph, Painting, Quickdraw, Real, and Sketch\}.

\subsection{Baselines} 
We compare POGM with the two most popular recent gradient-based DG, i.e., Fish \cite{2022-DG-Fish}, and Fishr \cite{2022-DG-Fishr}, using GPUs: NVIDIA RTX 3090.
Besides, we also compare our POGM with other state-of-the-art methods, which are ERM, IRM \cite{2022-DG-IRM}, GroupDRO \cite{2020-DG-GroupDRO}, Mixup \cite{2022-DG-CMixup}, DANN \cite{2023-DG-DANN}, MTL \cite{2021-DG-MTL}, SagNet \cite{2021-DG-SagNet}, ARM \cite{2021-DG-ARM}, VREx \cite{2021-DG-VREx}, RSC \cite{2020-DG-RSC}, AND-mask, SAND-mask \cite{2020-DG-SandMask}, EQRM \cite{2022-DG-EQRM}, RDM \cite{2024-DG-RDM}, SAGM \cite{2023-DG-SAGM}, CIRL \cite{2022-DG-CIRL}, MADG \cite{2024-FL-MADG}, ITTA \cite{2023-DG-ITTA}, MixStyle \cite{2021-DG-MixStyle}.

\subsection{Implementation Details}
We utilize different architectures for feature extraction and classification across datasets. Specifically, we employ a simple CNN for RMNIST and CMNIST, while ResNet-18 \cite{2016-DNN-ResNet} is used for VLCS, PACS, and OfficeHome. For Terra Incognita and DomainNet, ResNet-50 \cite{2016-DNN-ResNet} serves as the chosen architecture.
All experiments are trained for $100$ epochs. During the local domain training phase, we employ SGD to optimize both the feature extractor and classifier. The initial learning rates are set to $0.001$ for RMNIST and CMNIST, and $0.00005$ for PACS, VLCS, Terra Incognita, OfficeHome, and DomainNet. Batch sizes are set to 64 for RMNIST and CMNIST, and 32 for the other datasets.
In the meta-learning phase, the meta-learning rate is set to $0.01$, with the available searching hypersphere $\kappa$ set to $0.5$. We conduct $5$ local steps between each meta update.
Our code is based on DomainBed\footnote[1]{\url{https://github.com/facebookresearch/DomainBed/tree/main/domainbed}}.

\subsection{Hyperparameter Search}
Based on the experimental guidelines in \cite{2021-DG-DomainBed}, we perform a random search with 20 trials to fine-tune the hyperparameters for each algorithm and test domain. We divide the data from each domain into $80\%$ for training, and evaluation; and $20\%$ for selecting the hyperparameters. To mitigate randomness, we experiment twice with different seeds. Finally, we present the average results from these repetitions along with their estimated standard error.

We perform a grid search over pre-defined values for each hyperparameter and report the optimal values along with the values used for the grid search. Furthermore, we do early stopping based on the validation accuracy on the source domain and use the models that obtain the best validation accuracy.

\subsection{Model Selection} 
In DG, choosing the right model is like a learning task itself. We adopt the test-domain validation method from \cite{2021-DG-DomainBed}, which is one of three selection methods. This approach is like consulting an oracle, as we pick the model that performs best on a validation set with the same distribution as the test domain.

\section{Empirical Analysis}
\subsection{Detailed Measurement Setup of the Empirical Analysis}
\label{app:analysis-setup}
To evaluate the invariant gradient properties of Fish and Fishr, we conduct four following experiments.
\subsubsection{Domain-specific Model Norm Difference}\label{sec:domain-norm}
We conduct the domain-specific model norm difference to measure the empirical distance $\Vert\theta^{(r)}_i - \theta^{(r)}_{\Box}\Vert^2$ between domain-specific model $\theta^{(r)}_i$ and the learned model $\theta^{(r)}_{\Box}$, where $\Box = \{\textrm{Fish}, \textrm{Fishr}, \textrm{ERM}\}$ is the algorithms being evaluated.
To conduct the mentioned measurement, at each update round $r$, we conduct simultaneously two different training scenarios: 

\paragraph{Evaluated Algorithm Training.} From the previous model $\theta^{(r-1)}_{\Box}$, we train evaluated model $\theta^{(r)}_{\Box}$. 

\paragraph{Domain-specific Training.} From the previous model $\theta^{(r-1)}_{\Box}$, we train domain-specific model $\theta^{(r)}_i$ via the domain-specific data $S_i$. 

By doing so, we only evaluate the divergence between the evaluated and domain-specific models at every round. Thus, our metric guarantees fairness among algorithms.

\subsubsection{Domain-specific Gradient Angle}\label{sec:domain-angle}
We conduct the domain-specific model gradient angle to measure the cosine similarity $h^{(r)}_i \cdot h^{(r)}_{\Box} / \Vert h^{(r)}_i\Vert \Vert h^{(r)}_{\Box}\Vert$ between domain-specific gradient trajectory $h^{(r)}_i$ and the learned gradient trajectory $h^{(r)}_{\Box}$.
To conduct the mentioned measurement, at each update round $r$, we conduct simultaneously two different training scenarios: 

\paragraph{Evaluated Algorithm Training.} From the previous model $\theta^{(r-1)}_{\Box}$, we train evaluated model $\theta^{(r)}_{\Box}$. The evaluated gradient trajectory is measured as 
\begin{align}
    h^{(r)}_{\Box} = \theta^{(r)}_{\Box} - \theta^{(r-1)}_{\Box}.
\end{align}

\paragraph{Domain-specific Training.} From the previous model $\theta^{(r-1)}_{\Box}$, we train domain-specific model $\theta^{(r)}_i$ via the domain-specific data $S_i$. The domain-specific gradient trajectory is measured as
\begin{align}
    h^{(r)}_i = \theta^{(r)}_i - \theta^{(r-1)}_i.
\end{align}
By doing so, we only evaluate the divergence between the evaluated and domain-specific models at every round. Thus, our metric guarantees fairness among algorithms.

\subsubsection{Invariant Gradient Angle}\label{sec:inv-grad-angle}
To measure the invariant gradient angle, we first save a previously $\tau$-time trained model $\theta^{r-\tau}_{\Box}$. We measure the invariant gradient angle via the following cosine similarity formulation:
\begin{align}
    \textrm{Invariant Angle} = 
    \frac{[\theta^{(r)}_{\Box}- \theta^{(r-1)}_{\Box}] \cdot [\theta^{(r)}_{\Box}-\theta^{(r-\tau)}_{\Box}]}{\Big\Vert [\theta^{(r)}_{\Box}- \theta^{(r-1)}_{\Box}]\Big\Vert \times \Big\Vert  [\theta^{(r)}_{\Box}-\theta^{(r-\tau)}_{\Box}]\Big\Vert}
\end{align}
By doing so, we can approximate the fluctuation of the evaluated gradient. Specifically, as the gradient fluctuation issue is higher, the angle becomes larger, resulting in a smaller cosine similarity.
\subsection{Gradient Magnitude Norm}\label{sec:grad-norm}
We measure the gradient magnitude norm to evaluate how much the gradient made that round. As the gradient magnitude is larger, the gradient tends to progress more, and thus, make more impact on the learning progress. To measure the gradient magnitude norm, we calculate via the following formulation:
\begin{align}
   \textrm{Grad Norm} = \Vert\theta^{(r)}_{\Box} - \theta^{(r-1)}_{\Box}\Vert^2.
\end{align}

\clearpage
\subsection{Extensive Analysis on the Effect of Domain Divergence to the POGM Performance}\label{app:pogm-vs-domain-divergence}
\begin{figure}[!ht]
    \centering
    \includegraphics[width=0.5\textwidth]{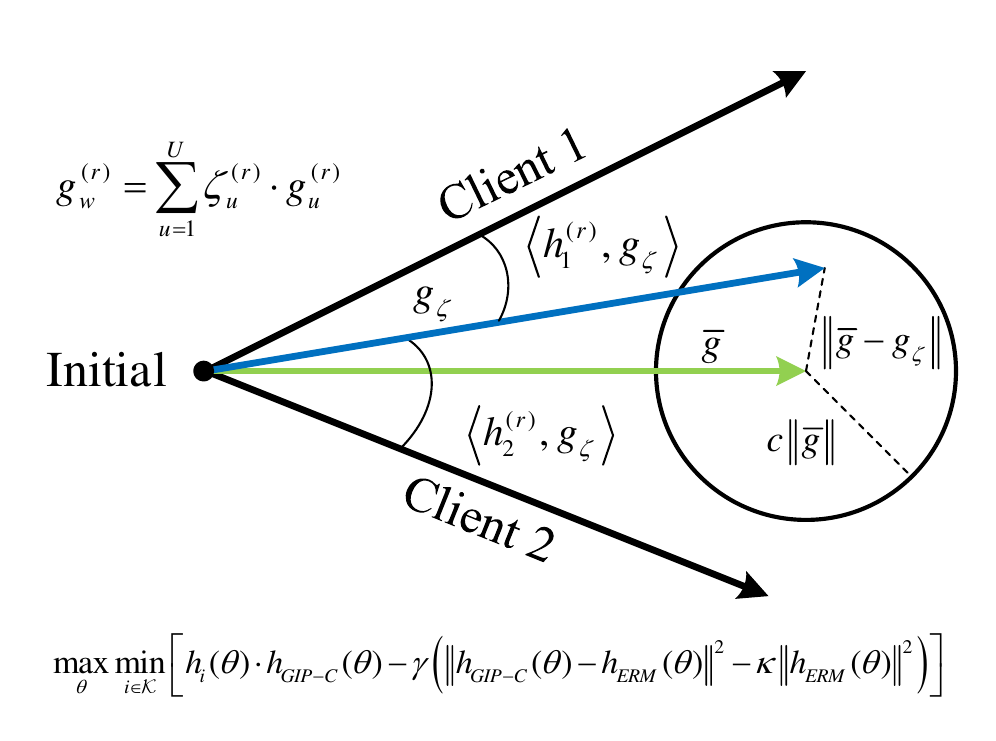}
    \caption{}
    \vspace{0.3cm}
    \label{fig:pogm-opt}
\end{figure}
To explain the effects of domain divergence on POGM performance, we first explain the rationale of POGM. As depicted in Fig.~\ref{fig:pogm-opt}, the POGM gradients are learned via the maximization of $\sum^{i\neq j}_{i,j\in\mathcal{K}}\nabla\mathcal{L}_i(\theta)\cdot\nabla\mathcal{L}_j(\theta)$. Therefore, the learned POGM gradient vectors will be between all domain-wise gradient trajectories, which can be found in Fig.~\ref{fig:pogm-opt}. 

When training domains are diverse, the space between them is extensive, increasing the chance of test domains to be. This increases the likelihood that the learned POGM gradients will align well with test domains, especially those similar to certain training domains. Conversely, when training domains are similar, and especially when test domains differ significantly from them, problems arise.
\begin{lemma}[Domain Divergence]
    Given a set of gradients vectors $\mathcal{G}_{\textrm{source}} = \{g_1, \ldots, g_K\}$ get from the model $\theta$ when train in source domains $\mathcal{K}=\{D-i\}^{K}_{i=1}$. If there is a gradient vector $g_L$ created by target domain $L$ such that $g_L\cdot g_i < g_i \cdot g_j,~\forall i,j\in K$, then $g_L$ resides outside the convex hull generated by the set $\mathcal{G}_{\textrm{source}}$.
\label{lemma:domain-divergence}
\end{lemma}
Lemma~\ref{lemma:domain-divergence} shows that, as the target domains diverge from the training source domains, the learned gradient of POGM in particular and via GIP, in general, will suffer from the divergence with that of the gradients on test target domains. Thus, we believe that, to guarantee the performance of the POGM, the training domain must diverge (and suggested to diverge than that of the test domains). Therefore, the POGM can learn a more generalized characteristics of source and target domains.

\subsection{Proof on Lemma~\ref{lemma:worst-case-grad}}
We have 
\begin{align}
    \nabla\mathcal{L}_i (\theta)\cdot \nabla\mathcal{L}_\textrm{GIP-C}
    \geq \min_{i\in\mathcal{K}} \nabla\mathcal{L}_i (\theta)\cdot \nabla\mathcal{L}_\textrm{GIP-C}.\notag
\end{align}
Therefore, we have:
\begin{align}
    \frac{1}{K}\sum^{}_{i\in\mathcal{K}}\nabla\mathcal{L}_i (\theta)\cdot \nabla\mathcal{L}_\textrm{GIP-C} 
    \geq \min_{i\in\mathcal{K}} \nabla\mathcal{L}_i (\theta)\cdot \nabla\mathcal{L}_\textrm{GIP-C}.\notag
\end{align}
\subsection{Proof on Lemma~\ref{lemma:GIP-C-Pareto}}
We have 
\begin{align}
    \theta^* &= \arg\max_{\theta} \sum^{}_{i\in\mathcal{K}}
    \mathcal{L}_i (\theta)\cdot \nabla\mathcal{L}_\textrm{GIP-C} (\theta) - \gamma\Big(\Vert \nabla\mathcal{L}_\textrm{GIP-C}(\theta) - \nabla\mathcal{L}_\textrm{ERM}(\theta)\Vert^2 - \kappa\Vert \nabla\mathcal{L}_\textrm{ERM}(\theta)\Vert^2\Big),
\end{align}
From Lemma~\ref{lemma:worst-case-grad} and definitions~\ref{def:pareto-dominance}, \ref{def:pareto-optimality}, we have: 
\begin{align}
    \mathcal{L}_{\textrm{avg}} (\theta^{*}) 
    & = \sum^{}_{i\in\mathcal{K}}
    \mathcal{L}_i (\theta^{*})\cdot \nabla\mathcal{L}_\textrm{GIP-C} (\theta^{*}) - \gamma\Big(\Vert \nabla\mathcal{L}_\textrm{GIP-C} (\theta^{*}) 
    - \nabla\mathcal{L}_\textrm{ERM} (\theta^{*})\Vert^2 - \kappa\Vert \nabla\mathcal{L}_\textrm{ERM} (\theta^{*})\Vert^2\Big) \notag \\
    & \geq 
     \min_{i\in\mathcal{K}} \Bigg[\mathcal{L}_i (\theta^{*})\cdot \nabla\mathcal{L}_\textrm{GIP-C} (\theta^{*}) - \gamma\Big(\Vert \nabla\mathcal{L}_\textrm{GIP-C} (\theta^{*}) 
    - \nabla\mathcal{L}_\textrm{ERM} (\theta^{*}) \Vert^2 
    - \kappa\Vert \nabla\mathcal{L}_\textrm{ERM} (\theta^{*})\Vert^2\Big)\Bigg] \notag\\
    & \geq \mathcal{L}_{\textrm{Pareto}} (\theta^{*}) = \mathcal{L}_{\textrm{Pareto}} (\theta^{*}). 
\end{align}
Therefore, we have that $\theta^{*}$ is also the Pareto optimality solution. 

\subsection{Proof on Lemma~\ref{lemma:domain-divergence}}
We define a convex hull $\mathcal{H}$ of the finite set $\{ g_1, \ldots, g_K \}$:
\begin{align}
    \mathcal{H} = \{h^{i},~\forall i\}= \{ \lambda_1 g_1 + \ldots + \lambda_K g_K ~\vert~\lambda
    \succeq 0, \mathbf{1}^{\top}\lambda = 1\},
\end{align}
it is obvious that to guarantee the point to be in the convex hull, we must satisfy the condition: $\lambda_k < 1,~\forall k$. Consider the target gradient vector $g_L$, where we have $g_L\cdot g_i < g_i \cdot g_j,~\forall i,j\in K$. To prove that $g_L$ is outside of the convex hull, we leverage a contradict consumption, where $g_L$ is in the convex hull, and  $g_L = \lambda_1 g_1 + \ldots + \lambda_K g_K$. Then, we have
\begin{align}
    g_L\cdot g_i 
    = (\lambda_1 g_1 + \ldots + \lambda_K g_K)\cdot g_i
    = \lambda_1 (g_1 \cdot g_i) + \ldots + \lambda_K (g_K \cdot g_i) 
    < (\lambda_1 + \ldots + \lambda_K)(g_L\cdot g_i) = g_L \cdot g_i
\end{align}
This does not exist, thus we have $g_L$ outside of the convex hull $\mathcal{H}$.

\subsection{Proof on Lemma~\ref{lemma:convergence}}
The update of the model can be written as follows: 
\begin{align}
    \theta^{(r,e+1)} = \theta^{(r,e)} - \eta \nabla U_k(\theta^{(r,e)}).
\end{align}
Now using the Lipschitz-smoothness assumption, we have
\begin{align}
    U_k(\theta^{(r,e+1)}) - U_k(\theta^{(r,e)}) 
   \leq -\eta\Big\langle \nabla U_k(\theta^{(r,e)}), \nabla U_k(\theta^{(r,e)})\Big\rangle + \frac{\eta^2 L}{2} \Big\Vert\nabla U_k(\theta^{(r,e)})\Big\Vert^2
=  \Big(\frac{\eta^2 L}{2} - \eta\Big) \Big\Vert \nabla U_k(\theta^{(r,e)})\Big\Vert^2.
\label{eq:1step-utility}
\end{align}
Averaging over all rounds, we have
\begin{align}
    U_k(\theta^{(r,e+1)}) - U_k(\theta^{(r,0)}) = \Big(\frac{\eta^2 L}{2} - \eta\Big)\sum^{E}_{e=0}\Big\Vert \nabla U_k(\theta^{(r,e)})\Big\Vert^2.
\end{align}
Rearranging terms, we have
\begin{align}
    \Big\Vert \nabla U_k(\theta^{(r,e)})\Big\Vert^2 \leq \frac{U_k(\theta^{(r,E^*)}) - U_k(\theta^{(r,0)})}{E^{*}\Big(\frac{\eta^2 L}{2} - \eta\Big)}.
\end{align}

\section{Proof on Theorems}
\subsection{Proof on Theorem~\ref{theorem:surrogate-IDGM}}
\begin{theorem}[Invariant Gradient Solution]
    Given the Pareto condition as mentioned in Lemma~\ref{lemma:GIP-C-Pareto}, $\widetilde{\pi} = \{\pi^{(r)}_{1},\ldots,\pi^{(r)}_{K}\}$ are the set of $K$ learnable scaling parameters for the joint learner at each $r$ communication round. The invariant gradient $h_\textrm{GIP-C}$ is characterized by the 
    \begin{align}
        h^{(r)}_\textrm{GIP-C} = h^{(r)}_{\textrm{ERM}} + \frac{\kappa\Vert h^{(r)}_{\textrm{ERM}}\Vert}{\Vert h^{(r)}_{\pi}\Vert}h^{(r)}_{\pi} \textrm{s.t.} 
        \quad 
        \widetilde{\pi} = \arg\min_{\pi} h^{(r)}_\pi\cdot h^{(r)}_{\textrm{ERM}} + \kappa\Vert h^{(r)}_{\textrm{ERM}}\Vert\Vert h^{(r)}_{\pi}\Vert
    \end{align}
    where $h^{(r)}_{\pi} = \sum^{K}_{i=1}\pi^{(r)}_i h^{(r)}_i$.
\end{theorem}

\textit{Proof.} 
For a clear proof, we denote $h^{(r)}_\textrm{GIP-C} = h^{(r)}_\textrm{GIP-C}(\theta)$, $h^{(r)}_\textrm{ERM} = h^{(r)}_\textrm{ERM}(\theta)$, $h^{(r)}_i = h^{(r)}_i(\theta)$, and $h^{(r)}_\pi = h^{(r)}_\pi (\theta)$. 
Therefore, we have
\begin{align}
     \max_{\theta}\min_{i\in\mathcal{K}}\Big[h^{(r)}_i (\theta)\cdot h^{(r)}_\textrm{GIP-C}(\theta) - \gamma\Big(\Vert h^{(r)}_\textrm{GIP-C}(\theta)-h^{(r)}_\textrm{ERM} (\theta)\Vert^2 - \kappa\Vert h^{(r)}_\textrm{ERM} (\theta)\Vert^2\Big)\Big].
\end{align}
This is equivalent to
\begin{align}
    \max_{\theta}\min_{\pi}\Big[h^{(r)}_\pi (\theta)\cdot h^{(r)}_\textrm{GIP-C}(\theta) - \gamma\Big(\Vert h^{(r)}_\textrm{GIP-C}(\theta)-h^{(r)}_\textrm{ERM} (\theta)\Vert^2 - \kappa\Vert h^{(r)}_\textrm{ERM} (\theta)\Vert^2\Big)\Big],
\end{align}
We first deal with the following minimization problem
\begin{align}
    \min_{\pi} U(\pi) = h^{(r)}_\pi (\theta)\cdot h^{(r)}_\textrm{GIP-C}(\theta) - \gamma\Big(\Vert h^{(r)}_\textrm{GIP-C}(\theta)-h^{(r)}_\textrm{ERM} (\theta)\Vert^2 - \kappa\Vert h^{(r)}_\textrm{ERM} (\theta)\Vert^2\Big).
\label{eq:minimization-problem}
\end{align}
To find the relaxation of the minimization $U(\pi)$, we fix $\pi, \gamma$ to find the optimal state of $h^{(r)}_\textrm{GIP-C}(\theta)$. The minimization is achieved when $\nabla_{h^{(r)}_\textrm{GIP-C}(\theta)} U(\pi) = 0$. For instance, we have:  
\begin{align}
    \nabla_{h^{(r)}_\textrm{GIP-C}(\theta)} U(\pi) 
    & = h^{(r)}_\pi (\theta) - 2\gamma \Big(h^{(r)}_\textrm{GIP-C}(\theta)-h^{(r)}_\textrm{ERM} (\theta)\Big)\nabla_{h^{(r)}_\textrm{GIP-C}(\theta)} (h^{(r)}_\textrm{GIP-C}(\theta)) \notag\\
    &= h^{(r)}_\pi (\theta) - 2\gamma \Big(h^{(r)}_\textrm{GIP-C}(\theta)-h^{(r)}_\textrm{ERM} (\theta)\Big) = 0.
\end{align}
This equality is achieved when 
\begin{align}
    h^{(r)}_\textrm{GIP-C}(\theta) &=  h^{(r)}_\textrm{ERM} (\theta) + \frac{h^{(r)}_\pi (\theta)}{2\gamma}.
\label{eq:approximate-solution}
\end{align}
Replace the solution found in \eqref{eq:approximate-solution} with \eqref{eq:minimization-problem}, we have: 
\begin{align}
    U(\pi) 
    &= h^{(r)}_\pi (\theta)\cdot \Big[h^{(r)}_\textrm{ERM} (\theta) + \frac{h^{(r)}_\pi (\theta)}{2\gamma}\Big] 
    - \gamma\Big(\Vert h^{(r)}_\textrm{GIP-C}(\theta)-h^{(r)}_\textrm{ERM} (\theta)\Vert^2 - \kappa\Vert h^{(r)}_\textrm{ERM} (\theta)\Vert^2\Big) \notag\\
    &= h^{(r)}_\pi (\theta)\cdot h^{(r)}_\textrm{ERM} (\theta) + h^{(r)}_\pi (\theta)\cdot\frac{h^{(r)}_\pi (\theta)}{2\gamma} 
    - \gamma\Big(\Vert h^{(r)}_\textrm{GIP-C}(\theta)-h^{(r)}_\textrm{ERM} (\theta)\Vert^2 - \kappa\Vert h^{(r)}_\textrm{ERM} (\theta)\Vert^2\Big) \notag\\
    &= h^{(r)}_\pi (\theta)\cdot h^{(r)}_\textrm{ERM} (\theta) + \frac{1}{2\gamma}\Vert h^{(r)}_\pi (\theta)\Vert^2 
    - \gamma\Big(\Vert \frac{h^{(r)}_\pi (\theta)}{2\gamma}\Vert^2 - \kappa\Vert h^{(r)}_\textrm{ERM} (\theta)\Vert^2\Big) \notag\\
    &= h^{(r)}_\pi (\theta)\cdot h^{(r)}_\textrm{ERM} (\theta) + \frac{1}{2\gamma}\Vert h^{(r)}_\pi (\theta)\Vert^2 
    - \frac{1}{4\gamma}\Vert h^{(r)}_\pi (\theta)\Vert^2 + \gamma\kappa\Vert h^{(r)}_\textrm{ERM} (\theta)\Vert^2 \notag\\
    &= h^{(r)}_\pi (\theta)\cdot h^{(r)}_\textrm{ERM} (\theta) + \frac{1}{4\gamma}\Vert h^{(r)}_\pi (\theta)\Vert^2
    + \gamma\kappa\Vert h^{(r)}_\textrm{ERM} (\theta)\Vert^2.
\label{eq:minimization-solution}
\end{align}
By fixing the $\gamma$, we have: 
\begin{align}
    \nabla_{\gamma} U(\pi) = 
    - \frac{1}{4\gamma^2}\Vert h^{(r)}_\pi (\theta)\Vert^2 + \kappa\Vert h^{(r)}_\textrm{ERM} (\theta)\Vert^2 = 0.
\end{align}
the optimal solution is when $\gamma$ satisfies: 
\begin{align}
    \gamma = \sqrt{\frac{1}{4}\Vert h^{(r)}_\pi (\theta)\Vert^2 \Big/ \kappa\Vert h^{(r)}_\textrm{ERM} (\theta)\Vert^2} = \frac{\Vert h^{(r)}_\pi (\theta)\Vert}{2\sqrt{\kappa}\Vert h^{(r)}_\textrm{ERM} (\theta)\Vert}.
\end{align}
Replace $\gamma$ into \eqref{eq:minimization-solution}, we have:
\begin{align}
    U(\pi) 
    &= h^{(r)}_\pi (\theta)\cdot h^{(r)}_\textrm{ERM} (\theta) + \sqrt{\kappa}\Vert h^{(r)}_\pi (\theta)\Vert \Vert h^{(r)}_\textrm{ERM} (\theta)\Vert.
\label{eq:minimization-solution-2}
\end{align}

\subsection{Proof on Theorem~\ref{theorem:invariant-property}}
Consider the GIP optimization problem from \eqref{eq:GIP-C-2}, taking the GIP value as $U_i=g_\textrm{GIP-C} \cdot g_i$, we consider the variance of the GIP function at each round $r$ (when each specific domain $k$ are being chosen to optimized) as:
\begin{align}
    \textrm{Var}\Big(U_i(\theta^{(r+1,e)}) \Big) 
    = \sum^{i\in K} U_i(\theta^{(r,e)}) - \max\min U_i(\theta^{(r,e)}) = \frac{K-1}{K} \Big(\sum^{i\in K}_{i\neq k} U_i(\theta^{(r,e)}) - \underset{\theta}{\max}~U_k(\theta^{(r,e)})\Big).
\label{eq:utility-variance}
\end{align}
After each step, only the $U_k$ is updated via maximization problem, i.e., $\theta^{(e+1)} = \theta^{(e)} - \eta \nabla U_k(\theta^{(e)})$. Then, we have:
\begin{align}
    U_k(\theta^{(r, e+1)}) 
    = U_k\Big(\theta^{(r, e)}- \eta \nabla U_k(\theta^{(r, e)})\Big) = U_k\Big(\theta^{(r, e)}\Big) - \eta \Big[\nabla U_k(\theta^{(r, e)})\Big]^2. 
\end{align}
Therefore, we have the \eqref{eq:utility-variance} as follows:
\begin{align}
    \textrm{Var}\Big(U_i(\theta^{(r+1,e)})\Big) = \frac{K-1}{K} \Big(\sum^{i\in K}_{i\neq k} U_i - U_k + \eta\Big[\nabla U_k(\theta^{(r, e)})\Big]^2\Big).
\label{eq:1step-utility-variance}
\end{align}
It is obvious that when $U_k(\theta^{(e+1)})$ conduct 1 step and in the next step, the different domain $k'$ is being chosen as Pareto fronts, we have the following:
\begin{align}
    \textrm{Var}\Big(U_i(\theta^{(r+1,e)})\Big)
    \leq \eta\Big\Vert\nabla U_k(\theta^{(r, e*)})\Big\Vert^2.
\label{eq:var-bound-1}
\end{align}
To bound the \eqref{eq:var-bound-1}, we have the following lemma: 
\begin{lemma}
    The gradient variance norm after $E^*$ rounds can be considered as
    \begin{align}
    \Big\Vert \nabla U_k(\theta^{(r,e)})\Big\Vert^2 
    \leq \frac{U_k(\theta^{(r,E^*)}) - U_k(\theta^{(r,0)})}{E^{*}\Big(\frac{\eta^2 L}{2} - \eta\Big)}.
    \end{align}
\label{lemma:convergence}
\end{lemma}
Therefore, we have: 
\begin{align}
    \textrm{Var}\Big(U_i(\theta^{(r+1,e)})\Big)
    \leq \frac{U_k(\theta^{(r,E^*)}) - U_k(\theta^{(r,0)})}{E^{*}\Big(\frac{\eta^2 L}{2} - \eta\Big)}
    = \frac{\textrm{Var}\Big(U_i(\theta^{(r+1,e)})\Big)}{E^{*}\Big(\frac{\eta^2 L}{2} - \eta\Big)}.
\label{eq:var-bound-2}
\end{align}

\subsection{Proof on Theorem~\ref{theorem:domain-convex-hull}}\label{appendix:domain-convex-hull}
From \cite{2022-DG-KLGuided}, we have
\begin{align}
    \LL_\textrm{test} 
    &\leq \LL_\textrm{train} + \frac{M}{2}\sqrt{\frac{1}{K}\sum^{K}_{i=1} D_\textrm{KL}\Big[p_T(y\vert \theta)\Vert p_i(y\vert\theta)\Big]}.
\end{align}
From Lemma~\ref{lemma:divergence-convex-hull}, we have: 
\begin{align}
    \LL_\textrm{test} 
    &\leq \LL_\textrm{train} + \frac{M}{2}\sqrt{\frac{1}{K^2}\sum^{K}_{i=1}\sum^{K}_{j=1} D_\textrm{KL}\Big[p_i(y\vert\theta)\Vert p_j(y\vert\theta)\Big]} \notag \\    
    &= \LL_\textrm{train} + \frac{M}{2}\sqrt{\frac{1}{K^2}\sum^{K}_{i=1}\sum^{K}_{j=1} D_\textrm{KL}\Big[p_i(y\vert x,\theta)p_i(x\vert\theta)\Vert p_j(y\vert x,\theta)p_j(x\vert\theta)\Big]} \notag\\
    &= \LL_\textrm{train} + \frac{M}{2}\sqrt{\underbrace{\frac{1}{K^2}\sum^{K}_{i=1}\sum^{K}_{j=1} D_\textrm{KL}\Big[p_i(y\vert x,\theta)\Vert p_j(y\vert x,\theta)\Big]}_{\textrm{B1}}
    + \underbrace{\frac{1}{K^2}\sum^{K}_{i=1}\sum^{K}_{j=1} D_\textrm{KL}\Big[p_i(x\vert\theta)\Vert p_j(x\vert\theta)\Big]}_{\textrm{B2}}},
\end{align}

where $\textrm{B1}$ is the divergence of the inference of model $\theta$ on different domains $i,j$. $\textrm{B2}$ is the divergence between domains, which is not tunable. This also means that, as the GIP $g_i\cdot g_j$ is maximized, we can improve the model generalization as follows:
\begin{align}
    \LL^{*}_\textrm{test} 
    &= \LL^{*}_\textrm{train} + 
    \frac{M}{2}\sqrt{\frac{1}{K^2}\sum^{K}_{i=1}\sum^{K}_{j=1} D_\textrm{KL}\Big[p_i(y\vert x,\theta^{*})\Vert p_j(y\vert x,\theta^{*})\Big]
    + \frac{1}{K^2}\sum^{K}_{i=1}\sum^{K}_{j=1} D_\textrm{KL}\Big[p_i(x\vert\theta^{*})\Vert p_j(x\vert\theta^{*})\Big]} \notag \\
    &\textrm{s.t.}~~~ \theta^{*} = \underset{\theta}{\arg\max} \sum^{i\neq j}_{i,j\in\mathcal{K}}\nabla\mathcal{L}_i(\theta)\cdot\nabla\mathcal{L}_j(\theta).
\end{align}

\end{document}